%% file: template.tex
\documentclass{article}

\usepackage{arxiv}

\usepackage[utf8]{inputenc} 
\usepackage[T1]{fontenc}    
\usepackage{hyperref}       
\usepackage{url}            
\usepackage{booktabs}       
\usepackage{amsfonts}       
\usepackage{nicefrac}       
\usepackage{microtype}      
\usepackage{lipsum}
\usepackage{graphicx}
\usepackage{amsmath}
\usepackage{amssymb}
\usepackage{tcolorbox}
\usepackage{soul}
\usepackage{amsthm}
\usepackage{csquotes}
\usepackage{bm}
\usepackage{lscape}

\usepackage{tikz}
\usetikzlibrary{decorations.text}
\usetikzlibrary{arrows.meta,bending,decorations.markings}

\hypersetup{
    colorlinks=true,
    linkcolor=blue,
    filecolor=blue,      
    urlcolor=blue,
}

\title{Graph representation forecasting of patient's medical conditions: towards a digital twin}

\author{
  Pietro~Barbiero\hspace{1mm}\href{https://orcid.org/0000-0003-3155-2564}{\includegraphics[scale=0.06]{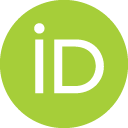}}\\
  Department of Computer Science and Technology\\
  University of Cambridge\\
  Cambridge, UK, CB3 0FD \\
  \texttt{pb737@cam.ac.uk} \\
   \And
 Ramon~Viñas~Torné\hspace{1mm}\href{https://orcid.org/0000-0003-2411-4478}{\includegraphics[scale=0.06]{orcid.png}}\\
  Department of Computer Science and Technology\\
  University of Cambridge\\
  Cambridge, UK, CB3 0FD \\
  \texttt{rv340@cam.ac.uk} \\
   \And
 Pietro~Li\'o\hspace{1mm}\href{https://orcid.org/0000-0002-0540-5053}{\includegraphics[scale=0.06]{orcid.png}}\\
  Department of Computer Science and Technology\\
  University of Cambridge\\
  Cambridge, UK, CB3 0FD \\
  \texttt{pl219@cam.ac.uk} \\
}

\begin{document}
\maketitle

\begin{abstract}
\textbf{Objective:} Modern medicine needs to shift from a wait and react, curative discipline to a preventative, interdisciplinary science aiming at providing personalised, systemic and precise treatment plans to patients. 
The aim of this work is to present how the integration of machine learning approaches with mechanistic computational modelling could yield a reliable infrastructure to run probabilistic simulations where the entire organism is considered as a whole.\\
\textbf{Methods:} We propose a general framework that composes advanced AI approaches and integrates mathematical modelling in order to provide a panoramic view over current and future physiological conditions.
The proposed architecture is based on a graph neural network (GNNs) forecasting clinically relevant endpoints (such as blood pressure) and a generative adversarial network (GANs) providing a proof of concept of transcriptomic integrability. \\
\textbf{Results:} We show the results of the investigation of pathological effects of overexpression of ACE2 across different signalling pathways in multiple tissues on cardiovascular functions. 
We provide a proof of concept of integrating a large set of composable clinical models using molecular data to drive local and global clinical parameters and derive future trajectories representing the evolution of the physiological state of the patient.\\
\textbf{Significance}: We argue that the graph representation of a computational patient has potential to solve important technological challenges in integrating multiscale computational modelling with AI. We believe that this work represents a step forward towards a healthcare digital twin.
\end{abstract}

\keywords{Precision medicine \and Graph representation \and Generative adversarial networks}

\section{Introduction}

Large part of humanity cannot afford basic medical care. Where  medicine has reached high level of biological and engineering technological complexity, costs have grown so much to reduce the access to a substantial percentage of the population. The target is very clear: we need healthcare to be precise and personalised at a reduced cost per capita in order to have 100$\%$ access \cite{tomavsev2020ai}.
A meaningful way to reach this target is to design a precise and  predictive medicine. Precision will require the integration of large amount of observations at individual and other observations  at population levels. These measures will be taken at different scales, from genome to clinical and family history and at systemic levels, i.e. considering multiple tissues and organs. We have no capacity to integrate such disparate information into equation-based models but we can use machine learning and, in particular, deep learning methods, to achieve this integration goal.
The predictive capability will allow us to shift from a wait and react, curative modus to a proactive modus of maintaining of the well-being condition or reverse disease status before more damage spreading (emergence of comorbidities and frailty).

Precision and predictable properties will therefore implicitly include personalised and preventative properties. The predictive capacity would act in lowering the overall and general cost of healthcare if the data integration step (leading to precision) would come at no extra cost. We believe that recent graph representation approaches \cite{Bronstein2017} could scale across all the variety of body signals at different levels and could sum all the above mentioned properties making possible a revolution in healthcare.

The development of reliable systems accurately forecasting physiological conditions of patients is one of the primary objectives of precision medicine \cite{noble2002rise}. The underlying complexity of human physiology makes this research field extremely challenging \cite{ginsburg2009genomic,naylor2010unraveling}. On one side, such complexity requires the development of models with very high capacity. On the other hand, the internal activity of several physiological systems are actually independent from others. For instance, the periodic contraction myocardiocytes does not depend on the pancreatic release of insulin. Nay, the mutual influence of several physiological processes is not hard-coded in their behaviour, but it is usually carried out by specialised transmitters, generating sophisticated signalling pathways. Besides, such signals can be transmitted from micro to macroscale systems and vice versa, affecting the whole organism at different levels \cite{wolkenhauer2014enabling,johnson2017enabling}. From a computational perspective, this means that each biological system could be modelled independently in the first place, provided that it receives expected signals from other systems \cite{chaplain2011multiscale}. 

This work proposes a new class of black-box machine-learning assisted methods for model analysis that scale to medical device deployment and run time monitoring and verification. By fusing ideas from systems medicine with scientific computing and machine learning we have developed a software that integrates and automates the analysis of vital parameters models under large uncertainty. A high degree of automation could transform how we use models in the scientific and medical discovery cycle and open up for a next-generation of powerful medical devices for probing the inner workings of full body in well-being and disease conditions.

The proposed architecture combines the qualities of a generative model \cite{goodfellow2014generative} with a graph-based representation \cite{scarselli2008graph} of pathophysiological conditions (see Figure \ref{fig:architecture}). On one side, the generative model can be used to produce synthetic data under different biological states that might not even be observed in reality. By augmenting the set of explorable states of the underlying biological system, the generative model may be employed for the simulation of extremely rare clinical scenarios representing precarious conditions which might be difficult to analyse otherwise \cite{yi2019generative}. In clinical contexts, this means that physicians will be able to set up personalised experiments in a virtual environment representing their patients in a very detailed and realistic way. On the other side, the graph model represents a virtual prototype of patients, a sort of "digital twin" mirroring the actual multiscale biological system \cite{gelernter1993mirror}, thus providing a general and flexible framework to run probabilistic simulations. The intrinsic characteristics of graph models make them suitable for the analysis of complex systems, while still providing highly interpretable results \cite{huang2020graphlime}. Graph are not just interpretable, but the network itself can be induced from data or even handcrafted. Researchers may take advantage of generative models for graphs to find optimal network configurations \cite{li2018learning} or formalise mathematical properties in form of differential equations or logical constraints with the purpose of describing the underlying system \cite{marra2019lyrics}. A panoramic view of individuals' conditions is provided by the final network configuration which combines information at organ, tissue, and cellular level. Cross-modal signals are also supported by the most recent graph learning frameworks, thus allowing the combination of different data sources, both structured and unstructured, real or simulated by generative methods. By relying upon flexible and modular architectures, our "digital twin" model can be conveniently deployed in dedicated hardware modules which might also be composed in a hierarchical fashion according to clinical needs.

\section{Design of a biomedical Digital Twin}
\label{sec:model}

The birth of the term "digital twin" could be the NASA's Apollo program where one spacecraft was launched into the outer space, while a "twin" spacecraft remained on earth to mirror flight conditions. Digital twin has been defined as “an integrated multiphysics, multiscale, probabilistic simulation of a vehicle or system that uses the best available physical models, sensor updates, fleet history, etc., to mirror the life of its flying twin” \cite{NASA, Grieves}. The Digital twin is a virtual prototype; the analysis of its digital life cycle provides information to understand a product’s functionality, manufacturing, behaviour and usage prior to building it.
Here the meaning of digital twin is slightly different: there is no product to be built, instead experimenting therapies on a digital twin will be cost effective and will provide us best practice to be used on the biological twin. 
Here an artificial intelligence model could enable the prediction of disease trajectories before the insurgence of symptoms. The personal medical digital twin could also represent a pragmatic way for the cyber-physical fusion, as a new approach to support biomedical engineering design. In our vision a composable AI architecture could enable the development of automatic analysis and verification techniques that are key to translational medicine.

This work proposes a modular approach which can be used to model the evolution of pathophysiological conditions.
The first module is based on a graph neural network (GNNs) forecasting clinically relevant endpoints (such as blood pressure) while the second one is represented by a generative adversarial network (GANs) providing a proof of concept of omic integrability. 
For experimental simulations we use and expand a clinical case study presented in \cite{barbiero2020computational} as a reference. In \cite{barbiero2020computational}, authors introduced a handcrafted dynamical system of equations representing some of the main physiological processes involved in cardiovascular and infection diseases. The computational model was developed to simulate systemic ripple effects caused by viral infections targeting the renin-angiotensin-aldosterone system (RAS) in patients with comorbid conditions. In such a context, the most relevant endpoints requiring constant monitoring include blood pressure, oxygenation, and insulin levels (for diabetic patients). In patients with multi-factorial diseases a very large set of factors may cause an irreparable impairment of such a delicate physiological equilibrium. Genomic traits, ageing, therapies, and lifestyle routines like physical exercise or dietary habits may all have an impact on keeping patients in a healthy state. Recording and keeping track of the stratification of all such factors is problematic by itself, but it is critical to get the whole picture. However, the actual challenge consists in exploiting the acquired information to forecast the evolution of clinical endpoints over time.
Depending on clinical needs and patient's comorbidities, it may be worth accounting for just a subset of such endpoints. Thus a sparse modelling approach whose components can be combined or activated on the fly may have a great potential in clinical practice.

\subsection{The effectiveness of GNNs and GANs in biomedical signal analysis}

Several properties of graph and generative adversarial neural networks make them suitable for medical data analysis:
\begin{itemize}
    \item Non-linearity: detection of non linear patterns is of key interest as most systems are inherently nonlinear in nature. Examples in medicine include heart rate dynamics, pulmonary functions, vascular structure, and gait dynamics. There is often a loss of non linearity and multiscale fractal in ageing and disease conditions \cite{goldberger2002fractal}.
    \item Interpretability: the possibility of interpreting the behaviour of models and the reason for their predictions is pivotal if not critical for in many fields including clinical practice. Thanks to their structure, graph-based models are much easier to interpret with respect to other neural approaches.
    \item Non-Euclidean geometry: tissue and organ distributions could be modelled as graph models where each node or the graph contain time-dependent signals; similarly for  pressure and electric sensors positioned at various parts of the body.
    Lymphatic vessels can also be modelled as a network where lymph nodes are vertices. At lower scale, cell arrangements in tissues form particular manifolds; proteins and genes are organised in regulatory networks; other examples are cytoskeleton and organelles (mitochondria networks). Additionally, diseases could be seen as nodes in a graph where edges represent comorbidity or underlying polygenic causes.  
    \item Modularity: A key property of GNNs is modularity, which allows to learn independent mechanisms that can be reused in several parts of the graph. Modularity facilitates scalability and allows to model dynamic properties of graphs.
    \item Cross-modality: both GNNs and GANs can learn how to combine structured and unstructured data sources, spanning different levels of biological complexity. This is particularly relevant when integrating signals at different levels of biological scale such as DNA methylation and fMRI data.
    \item Generative: both GNNs and GANs can learn how to generate new data preserving the statistical properties of the training set. This could be used to compare statistics at individual level with those at specific groups identified with stratification analysis or at general population levels. 
    \item Multiscale: the graph representation has the capability of integrating granular information organised as networks at different layers of biological complexity. This allows to recognise patterns in higher-order structures such as motifs, pathways, tissues (as compositions of cells), organs (as composition of tissues), processes and apparatus (as composition of organs), stratification (as composition of individuals). 
    \item Spectral density: together with spatial properties, GNN are amenable to frequency domain analysis. This allows to investigate network motifs, substructures and periodical patterns at network levels.
\end{itemize}

\subsection{Graph neural model} \label{sec:graph_nn}

Graphs are mathematical structures which are used to model a set of objects (nodes) and their mutual relationships (edges) \cite{bollobas2013modern}. Graphs are employed in a variety of research areas as they provide a general and flexible data structure for modelling real-world systems \cite{zhou2018graph,lieberman2005evolutionary,rakocevic2019fast,bica2020unsupervised}.
Graph neural networks (GNN) are deep learning-based models working on the graph domain \cite{scarselli2008graph,battaglia2018relational,wu2020comprehensive}.
Their properties have been recently drawn the attention of the artificial intelligence research community given their high interpretability and as the only non-Euclidean models available in machine learning \cite{lecue2019role,huang2020graphlime}. The combination of graph theory and neural network elements have made GNNs one of the most promising tools to analyse complex systems in the graph domain. From neural networks GNNs inherit a data-driven approach associated with a multi-layer architecture, which is the key to extract hierarchical patterns from data. However, unlike other deep-learning models, GNNs exploit additional features from graph theory and other mathematical disciplines. The main advantage with respect to other machine learning models relies in their extremely flexible and interpretable architecture. Once defined, the main endpoints of a system together with their mutual relationships directly induce a corresponding graph representation, which can be easily interpreted from a human standpoint. The abstract graph representation can be handcrafted, when the complexity of the underlying system allows it, or even automatically induced from data using generative approaches \cite{li2018learning}. Hybrid techniques may also be explored taking advantage of generative algorithms for handling system complexity and human modelling to customise the most relevant endpoints. The design of GNNs is based on two basic principles, flexibility and composability. GNNs support different graph structures as well as flexible representations of global, node, and edge attributes, customizable according to specific demands of tasks.

\input{architecture}

\subsubsection{Graph network blocks}
The GNN framework proposed by Battaglia et al.  \cite{battaglia2018relational} is based on modules called graph network blocks (GN blocks) representing the core computation units of a GNN. Multiple GN blocks can be composed or even combined with other neural networks to generate complex architectures. 
A graph neural network can be defined as a 3-tuple $G = (\textbf{u}, H, E)$. $H = \{\textbf{h}_i\}_{i=1:N^v}$ is the node set where the feature of each node is denoted by $\textbf{h}_i$. $E = \{ ( \textbf{e}_k, r_k, s_k ) \}$ is the edge set where each node is represented by its features $\textbf{e}_k$, the receiver node $r_k$, and the sender node $s_k$. \textbf{u} denotes a set of global attributes representing the state of the underlying system. Each GN block consists of three update functions, $\phi$, and three aggregation functions, $\rho$:

\begin{eqnarray}
    \label{eq:gnn}
    &\textbf{e}_k'& = \phi^e (\textbf{e}_k, \textbf{h}_{r_k}, \textbf{h}_{s_k}, \textbf{u}) \qquad\qquad \bar{\textbf{e}}_i' = \rho^{e \rightarrow h} (E_i') \nonumber \\
    &\textbf{h}_i'& = \phi^h (\bar{\textbf{e}}_k, \textbf{h}_i, \textbf{u}) \qquad\qquad\qquad\: \, \bar{\textbf{e}}' = \rho^{e \rightarrow u} (E') \\
    &\textbf{u}'& = \phi^u (\textbf{e}', \textbf{h}', \textbf{u}) \qquad\qquad\qquad\; \: \bar{\textbf{h}}' = \rho^{h \rightarrow u} (H') \nonumber
\end{eqnarray}

where $E_i' = \{ ( \textbf{e}_k', r_k, s_k ) \}$, $H' = \{ (\textbf{h}_i') \}_{i=1:N^v}$, and $E' = \bigcup_i E_i' = \{ ( \textbf{e}_k', r_k, s_k ) \}_{k=1:N^e}$.
In order to train a GN block in full, six computation steps are required, alternating the update and aggregation functions. For each edge, $E'_i$ is computed through the update function $\phi^e$. The result is then aggregated by means of the function $\rho^{e \rightarrow v}$. The output $\bar{\textbf{e}}_i'$ corresponds to an edge update and it is employed to update node representations $\textbf{h}_i'$ by means of $\phi^h$. $\rho^{e \rightarrow u}$ and $\rho^{h \rightarrow u}$ perform aggregation steps generating $\bar{\textbf{e}}'$ and $\bar{\textbf{h}}'$ from edge and node updates, respectively. Global attributes represented by $\textbf{u}'$ are computed leveraging the information from $\bar{\textbf{e}}'$, $\bar{\textbf{h}}'$, and $\textbf{u}$ via the function $\phi^u$.
The learning process of each GN block may be independent or co-dependent with other blocks. Constraints may apply on edges, information flows, or global attributes, depending on the application. In this work we are just interested in the evaluation of global attributes to monitor clinical endpoints and we did not apply any learning constraint, even if in clinical practice may still be of great interest. Given a set of labels for global attributes $\textbf{t} = \{t_i\}_{i=1:N^v}$ and the corresponding predictions provided by the GN block $\widehat{\textbf{u}}' = \{\widehat{u}_i'\}_{i=1:N^v}$ representing the evolution of the underlying biological system, we aim at minimising the following objective function:

\begin{equation}
    \min_\theta \sum_{i=1}^{N^v} \big(t_i - \widehat{u}_i'\big)
\end{equation}

where $\theta$ is the set of model's parameters.

\subsubsection{Graph layers}

GNNs natively allow the design of complex systems using a modular approach. First, the complexity is broken up by developing independent subsystems representing genomic alterations, biological pathways, and organ physiology. Each subsystem can be represented as a different node or set of nodes in a GNN, while inter-process signals can be reframed as message passing operations supporting multiscale ripple effects. Homogeneous subsystems can be aggregated into layers according to their characteristics.
Our digital patient model is composed of four layers: the transcriptomic layer, the cellular layer, the organ layer, and the exposomic layer. Other layers that bring information on other omics or body sensor network i.e. a collection of networked sensors that can be used to monitor human physiological signals, could be similarly implemented.

\paragraph{Transcriptomic layer}
The transcriptomic layer operates on the set of RNA transcripts produced by the genome at a particular time. Currently, RNA sequencing (RNA-seq) can measure RNA abundance across the entire genome with high resolution. The resulting high-throughput gene expression data can be used to uncover disease mechanisms \cite{cookson2009mapping,emilsson2008genetics,gamazon2018using}, propose novel drug targets \cite{sirota2011discovery,evans2004moving}, provide a basis for comparative genomics \cite{colbran2019inferred}, and address a wide range of fundamental biological problems.  

In this work, we study the crosstalk between tissues in the organ layer (see Figure \ref{fig:architecture}) through the communicome, e.g. communication factors in blood \cite{Ray2007}. Specifically, we analyse to what extent the expression of genes involved in the renin-angiotensin system can be explained by genes from signalling and receptor pathways, including the chemokine, TNF, and TGF-$\beta$ pathways. We further develop a transcriptomics generative model based on a generative adversarial network \cite{gan} and simulate the effects of SARS-CoV-2 infection by conditioning on high expression of ACE2 in the lung, kidney, and pancreas.

\paragraph{Cellular layer}
The cellular layer involves biological processes affecting individual cells from metabolism and protein synthesis to replication and motility. In this study we focus on modelling the Renin-Angiotensin System (RAS), one of the main biological pathways regulating blood pressure and closely related to SARS-CoV-2 infectivity. Hence, it represents a suitable case study to demonstrate the flexibility and expressiveness of our GNN-based approach.
The renin-angiotensin system is a hormone system regulating vasoconstriction and inflammatory response \cite{fountain2019physiology}. The key hormone of the system is the peptide Angiotensin II (ANG-II) generated from the decapeptide Angiotensin I by the angiotensin-converting enzyme (ACE). ANG II promotes vasoconstriction, hypertension, inflammation, and fibrosis by activating the ANG-II type 1 receptor (AT1R) \cite{kuba2010trilogy,gironacci2011angiotensin}.
Glucose concentration, ACE inhibitor treatments, and viral infections binding to ACE2, such as SARS-CoV-2, can all have a significant impact on the RAS. A high glucose concentration may determine chronic hypertensive conditions. Reducing ANG II production with ACE inhibitors increases vasodilation and vasoprotection effects stimulated by the overproduction of AT2R and ANG-(1-7) \cite{zaman2002drugs}. Viral infections such as SARS-CoV-2 may also have an impact on RAS, as the virus binds to ACE2 in order to gain entry into the host cell. This results in an altered ACE2 activity and concentration, possibly leading to hypertension and inflammatory response \cite{south2020controversies}.

\paragraph{Organ layer}
The organ layer comprises group of tissues with similar functions (organs) and complex networks of cooperating organs. Given the nature of the multi-factorial disease under study, we limited the organ layer to the circulatory system and a physiological representation of a few organs \cite{barbiero2020computational}: heart, lungs, and kidneys.
The heart model includes four compartments known as chambers \cite{neal2007subject}. Deoxygenated blood collected from the superior and inferior venae cavae flows into the right atrium. When the right atrium contracts, the blood is pumped through the tricuspid valve into the right ventricle. From the right ventricle the blood is pumped into the pulmonary trunk through the pulmonary valve flowing towards the lungs where carbon dioxide is exchanged for oxygen. The pulmonary circulation is composed of five vascular segments: proximal and distal pulmonary artery, small arteries, capillaries, and veins. Oxygenated blood collects into the left atrium via the pulmonary veins. From there, it flows into the left ventricle through the mitral valve and it is pumped into the aorta through the aortic valve for systemic circulation, providing oxygen and nutrients to body cells for metabolism in exchange for carbon dioxide and waste products.
The mean arterial blood pressure is controlled by baroreceptors, special sensory neurons excited by a stretch in the carotid sinus and aortic arch vessels. They relay sensory information regarding blood pressure changes to the central nervous system where it is processed and utilised primarily in autonomic reflexes, regulating short-term blood pressure.

\paragraph{Exposomic layer}
The exposome refers to the totality of exposure individuals experience from conception until death and its impact on chronic and acute diseases \cite{wild2005complementing}. Toxicants, dietary regimens, treatments, physical exercise, posture, lifestyle habits, all of them are possible exposures taking part to individual's well-being or disease condition. All such environmental factors are deeply coupled among themselves but also with individuals influencing the effects of new or present exposures. The exposome is intrinsically co-dependent on a person's genetics, epigenetics, health status, and physiology. For instance, regular exposure to pollution may lead to the outbreak of a lung carcinoma which in turn may call for clinical intervention. In this work, we consider four types of exposures: dietary habits, physical activity, therapeutic treatments, and viral infections.

\subsubsection{Inter-process signals and clinical endpoints}
One of the main advantages of using GNN-based models relies in that inter-process and multiscale communications can be natively implemented using message passing. In a GNN, each biological entity can be represented as a node while the relationship between two entities can be modeled using directional edges. Signals exchanged between nodes are implemented using message functions $\phi^h$ (see Eq. \ref{eq:gnn}) which are used to update the hidden states of nodes. Such state transition will then have an impact on messages exchanged at the following time steps.
Another strength of GNN models consists in the possibility of supervising the evolution of the underlying system, by using the readout
functions $\phi^u$. Hence, the endpoints of multi-factorial diseases can be directly controlled by checking the output of readout functions in critical nodes.
The resulting GNN model will combine a simple and modular design with a versatile structure accommodating for complex multiscale systems where clinical endpoints can be easily monitored and forecast in real time.

\subsubsection{Assessing prediction uncertainty}
\label{sec:uncertainty}
The aim of developing a digital patient model is to provide an accurate estimation of the trajectory of a patient by forecasting clinically relevant endpoints. In such a context, quantifying model uncertainty is critical. One of the most established techniques relies upon the use of dropout \cite{srivastava2014dropout} at test time, as a Bayesian approximation, without sacrificing either computational complexity or test performance \cite{gal2016dropout}. In this framework, the first two moments of the predictive  distribution $q$ performing $T$ stochastic forward passes for a sample $\textbf{x}^*$ with label $\textbf{y}^*$ can be estimated as \cite{gal1506dropout}:

\begin{equation}
    \mathbb{E}_{q(\textbf{y}^*,\textbf{x}^*)} (\textbf{y}^*) \approx \frac{1}{T} \sum_{i=1}^T \widehat{\textbf{y}}^* (\textbf{x}^*, W_1^t, \dots, W_L^t)
\end{equation}

\begin{eqnarray}
    \textrm{Var}_{q(\textbf{y}^*,\textbf{x}^*)} (\textbf{y}^*) \approx \tau^{-1} I_D &+& \frac{1}{T} \sum_{i=1}^T \widehat{\textbf{y}}^* (\textbf{x}^*, W_1^t, \dots, W_L^t)^T \widehat{\textbf{y}}^* (\textbf{x}^*, W_1^t, \dots, W_L^t) \nonumber \\
    &-& \mathbb{E}_{q(\textbf{y}^*,\textbf{x}^*)} (\textbf{y}^*)^T \mathbb{E}_{q(\textbf{y}^*,\textbf{x}^*)} (\textbf{y}^*)
\end{eqnarray}

where $\widehat{\textbf{y}}^*$ is the predicted label, $\{W_i\}_{i=1}^L$ is a set of random variables representing the weights of a neural network with $L$ layers, $I_D$ is an identity matrix, $D$ is the number of output units of the neural network, and $\tau$ is a precision hyper-parameter. The method has also been generalised to convolutional \cite{gal2015bayesian} and recurrent networks \cite{gal2016theoretically}.

Here we show how such technique can be used to quantify the uncertainty of a GNN by generating a predictive distribution of the trajectories representing the future states of the patient.
Let $x_1^*, \dots, x_k^*$ be a sequence of real values representing a clinical endpoint measured at $1,\dots,k$ time steps. Let $f^t$ be a stochastic model which takes a sequence $x_1^*, \dots, x_k^*$ as input and it outputs a prediction $\widehat{y}^* \in \mathbb{R}$. We are interested in estimating a predictive distribution of the trajectories of the variable $x$ over the next $k+1,\dots,k+h$ time steps. To this aim, we can use an iterative algorithm by generating one trajectory at a time. The first prediction $\widehat{y}_{k+1}^*$ can be generated as:

\begin{equation}
    \widehat{y}_{k+1}^{*,t} = f^t (x_1^*, \dots, x_k^*)
\end{equation}

By using the obtained prediction and sliding the time window one time step further, we can generate the first prediction for the second time step $k+2$:

\begin{equation}
    \widehat{y}_{k+2}^{*,t} = f^t (x_2^*, \dots, x_k^*, \widehat{y}_{k+1}^{*,t})
\end{equation}

The procedure can be repeated for $k+h$ time steps to generate a single trajectory. Model uncertainty can be assessed building multiple trajectories by performing $T$ stochastic forward passes. The resulting algorithm is equivalent to a Monte Carlo sampling as proven by Gal et al. \cite{gal2016dropout}. In our GNN model, the approach we just described can be easily applied for each node in order to assess the uncertainty of clinical endpoints.

\subsection{Generative adversarial model}\label{sec:gan_generative}

One way of studying probability distributions is by means of generative models, which describe the random phenomenon in terms of the joint probability distribution of observed and target variables \cite{jebara2012machine}. Generative adversarial networks (GANs) are a framework for estimating generative models via an adversarial process \cite{gan}. They are often described as a two-player game in which both players are encouraged to improve. One player, the \emph{generator}, creates samples that are intended to be indistinguishable from the ones coming from a given data distribution. The other player, the \emph{discriminator}, learns to determine whether samples come from the \emph{fake} distribution (\emph{fake} samples) or the \emph{real} data distribution (\emph{real} samples). Figure \ref{gan} shows the basic idea of generative adversarial networks. With respect to other generative models, they provide a general and flexible framework for the analysis of joint probability distributions. The architecture itself allows a fine control of the data generation process and a high level of customisation, making them suitable for a variety of experimental scenarios.

\input{gan_framework}


\subsubsection{Crosstalk between tissue-types}
The activity of biological systems is determined by internal factors, determined by intrinsic and functional properties, and by external factors shaping the interconnections between different systems. Chemical and molecular events, like oxygenation or protein phosphorylation, are often the vehicles of biological signals' transduction. A chain of biochemical events forms a signalling pathway whose activation may give rise to a biochemical cascade of events affecting the organism at different levels. In complex organisms several signal transduction pathways communicate and react reciprocally generating biological crosstalks. Crosstalks have been widely characterised and observed in a variety of biological processes from micro to macroscale from genomics \cite{du2015dna,poyton1996crosstalk}, to internal and external cell activity \cite{li2016crosstalk,geiger2001transmembrane}, and even between tissues \cite{lengyel2018cancer}.
Here, we develop a generative model based on a generative adversarial network to produce synthetic transcriptomics data describing the ripple effects of a viral infection on crosstalks between different tissues. 
The aim is to demonstrate how generative approaches can be used both to reproduce and enhance the set of observable states of a patient allowing for a deeper understanding of complex biological processes.

\textbf{Model. } Consider a dataset $\mathcal{D} = \{(\mathbf{x}, \mathbf{m}, \mathbf{r}, \mathbf{q})\}$ of samples from an unknown distribution $\mathbb{P}_{\mathbf{x}, \mathbf{m}, \mathbf{r}, \mathbf{q}}$, where $\mathbf{x} \in \mathbb{R}^{t \times n}$ represents a vector of a patient's gene expression values in $t$ tissues; $n$ is the number of genes; $\mathbf{m} \in \{0, 1\}^t$ is a mask vector indicating whether the expression of each tissue has been measured for the given patient; and $\mathbf{r} \in \mathbb{R}^k$ and $\mathbf{q} \in \mathbb{N}^c$ are vectors of $k$ quantitative covariates (e.g. age) and $c$ categorical (e.g. gender), respectively. Our goal is to produce realistic gene expression samples by modelling the conditional probability distribution $\mathbb{P}(\mathbf{X}=\mathbf{x} | \mathbf{M}=\mathbf{m}, \mathbf{R}=\mathbf{r}, \mathbf{Q}=\mathbf{q})$, where $\mathbf{r}$ includes the expression of ACE2 in different tissues (e.g. lung, kidney, and pancreas). By modelling this distribution, we can sample data for different conditions and quantify the uncertainty of the generated expression values.

Our method builds on a Wasserstein GAN with gradient penalty (WGAN-GP) \cite{wgan, wgangp}. Similar to Generative Adversarial Networks (GAN) \cite{gan}, WGAN-GPs estimate a generative model via an adversarial process driven by the competition between two players, the \emph{generator} and the \emph{critic}.

The generator aims at producing samples from the conditional $\mathbb{P}(\mathbf{X}| \mathbf{M}, \mathbf{R},  \mathbf{Q})$. Formally, we define the generator as a function $G_{\theta}: \mathbb{R}^u \times \mathbb{R}^k \times \mathbb{N}^c \rightarrow \mathbb{R}^{t \times n}$ parametrised by $\theta$ that generates gene expression values  $\hat{\mathbf{x}}$ as follows:
\begin{equation}\label{eq:gen}
    \hat{\mathbf{x}} = \mathbf{m} \odot G_{\theta}(\mathbf{z},  \mathbf{r}, \mathbf{q})
\end{equation}
where $\mathbf{z} \in \mathbb{R}^u$ is a vector sampled from a fixed noise distribution $\mathbb{P}_\mathbf{z}$ and $u$ is a user-definable hyperparameter. We apply the mask $\mathbf{m}$ element-wise to match the distribution of missing tissues of the training dataset.


The critic takes gene expression samples $\mathbf{x}$ from two input streams (the generator and the data distribution) and attempts to distinguish the true input source. Formally, the critic is a function $D_{\omega}: \mathbb{R}^{t \times n} \times \{0, 1\}^t \times \mathbb{R}^k \times \mathbb{N}^c \rightarrow \mathbb{R}$ parametrised by $\omega$ that we define as follows:
\begin{equation*}
    \bar{y} = D_{\omega}(\bar{\mathbf{x}}, \mathbf{m}, \mathbf{r}, \mathbf{q})
\end{equation*}
where the output $\bar{y}$ is an unbounded scalar that quantifies the degree of realism of an input sample $\bar{\mathbf{x}}$ given the covariates $\mathbf{r}$ and $\mathbf{q}$ (e.g. high values correspond to real samples and low values correspond to fake samples). When the expression of a certain tissue is unavailable for a given patient, we impute the unobserved values of $\bar{\mathbf{x}}$ with zeroes and set the corresponding value of the mask $\mathbf{m}$ to zero.

We optimise the generator and the critic adversarially. Following \cite{wgan}, we train the generator $G_{\theta}$ and the critic $D_{\omega}$ to solve the following minimax game based on the Wasserstein distance:
\begin{align}\label{eq:minimax}
\begin{split}
    \min_{\theta} \max_{\omega}
    &\mathop{\mathbb{E}}_{\mathbf{x}, \mathbf{m}, \mathbf{r}, \mathbf{q} \sim \mathbb{P}_{\mathbf{x}, \mathbf{r}, \mathbf{q}}}\Big[D_{\omega}(\mathbf{x}, \mathbf{m}, \mathbf{r}, \mathbf{q}) -  \mathop{\mathbb{E}}_{\mathbf{z} \sim \mathbb{P}_{\mathbf{z}}}
    [D_{\omega} (\hat{\mathbf{x}}, \mathbf{m}, \mathbf{r}, \mathbf{q})] \Big] \\
    \qquad \text{subject to} &\qquad || D_{\omega}(\mathbf{x}_i, \mathbf{m}, \mathbf{r}, \mathbf{q}) -  D_{\omega}(\mathbf{x}_j, \mathbf{m}, \mathbf{r}, \mathbf{q}) || \le || \mathbf{x}_i - \mathbf{x}_j || \\ 
    &\qquad \forall \mathbf{x}_i, \mathbf{x}_j \in \mathbb{R}^{t \times n},
    \mathbf{m}  \in \{0, 1\}^t,
    \mathbf{r} \in \mathbb{R}^k, \mathbf{q} \in \mathbb{N}^c
\end{split}
\end{align} 
where $\hat{\mathbf{x}}$ is defined as in Equation \ref{eq:gen} and the constraint enforces the critic $D_{\omega}$ to be 1-Lipschitz, that is, the norm of the critic's gradient with respect to $\mathbf{x}$ must be at most 1 everywhere.


Let $\{(\mathbf{x}_i, \mathbf{m}_i, \mathbf{r}_i, \mathbf{q}_i)\}_{i=1}^{b}$ be a mini-batch of $b$ independent samples from the training dataset $\mathcal{D}$. Let $\{\mathbf{z}_1, \mathbf{z}_2, ..., \mathbf{z}_k\}$ be a set of $k$  vectors sampled independently from the noise distribution $\mathbb{P}_{\mathbf{z}}$ and let us define the synthetic samples corresponding to the mini-batch as $\hat{\mathbf{x}}_i = \mathbf{m}_i \odot G_{\theta}(\mathbf{z}_i, \mathbf{r}_i, \mathbf{q}_i)$  for each $i$ in $[1, 2, ..., k]$. We solve the minimax problem described in Equation \ref{eq:minimax} by interleaving mini-batch gradient updates for the generator and the critic, optimising the following problems:
\begin{align}\label{eq:minimax_impl}
    \begin{split}
        \text{Generator: } \quad \min_{\theta} \quad 
        -&\frac{1}{k}\sum_{i=1}^k D_{\omega} \big(\hat{\mathbf{x}}_i, \mathbf{m}_i, \mathbf{r}_i, \mathbf{q}_i\big) \\
        \text{Critic: } \quad \min_{\omega} \qquad 
        \ &\frac{1}{k}\sum_{i=1}^k D_{\omega} \big(\hat{\mathbf{x}}_i, \mathbf{m}_i, \mathbf{r}_i, \mathbf{q}_i\big) - D_{\omega} (\mathbf{x}_i, \mathbf{m}_i, \mathbf{r}_i, \mathbf{q}_i) \\
        + & \frac{\lambda}{k} \sum_{i=1}^k \big(||\nabla_{\tilde{\mathbf{x}}_i} D_{\omega} (\tilde{\mathbf{x}}_i, \mathbf{m}_i, \mathbf{r}_i, \mathbf{q}_i)||_2 - 1\big)^2
    \end{split}
\end{align}
where $\lambda$ is a user-definable hyperparameter and each $\tilde{\mathbf{x}}_i$ is a random point along the straight line that connects $\mathbf{x}_i$ and $\hat{\mathbf{x}}_i$, that is, $\tilde{\mathbf{x}}_i = \alpha_i \mathbf{x}_i + (1 - \alpha_i) \hat{\mathbf{x}}_i$ with $\alpha_i \sim \mathcal{U}(0, 1)$. Intuitively, since enforcing the 1-Lipschitz constraint everywhere is intractable (see Equation \ref{eq:minimax}), the second term of the critic problem is a relaxed version of the constraint that penalises the gradient norm along points in the straight lines that connect real and synthetic samples \cite{wgangp}.

\textbf{Architecture. } Figure \ref{fig:architecture} shows the architecture of both players.  The generator $G$ receives a noise vector $\mathbf{z}$ as input (green box) as well as sample covariates $\mathbf{r}$ and $\mathbf{q}$ (orange boxes) and produces a vector $\hat{\mathbf{x}}$ of synthetic expression values (red box). The critic $D$ takes either a real gene expression sample $\mathbf{x}$ (blue box) or a synthetic sample $\hat{\mathbf{x}}$ (red box), in addition to sample covariates $\mathbf{r}$ and $\mathbf{q}$, and attempts to distinguish whether the input sample is real or fake. For both players, we use word embeddings \cite{wordembeddings} to model the sample covariates (light green boxes), a distinctive feature that allows to learn distributed, dense representations for the different tissue types and, more generally, for all the categorical covariates $\mathbf{q} \in \mathbb{N}^c$.

Formally, let $q_j$ be a categorical covariate (e.g. tissue type) with vocabulary size $v_j$, that is, $q_j \in \{1, 2, ..., v_j \}$, where each value in the vocabulary $\{1, 2, ..., v_j \}$ represents a different category (e.g. whole blood or kidney). Let $\bar{\mathbf{q}}_j \in \{0, 1\}^{v_j}$ be a one-hot vector such that $\bar{q}_{jk}=1$ if $q_j = k$ and $\bar{q}_{jk}=0$ otherwise. Let $d_j$ be the dimensionality of the embeddings for covariate $j$. We obtain a vector of embeddings $\mathbf{e}_j \in \mathbb{R}
^{d_j}$ as follows:
\begin{equation*}
    \mathbf{e}_j = \mathbf{W}_j  \bar{\mathbf{q}}_j
\end{equation*}
where each $\mathbf{W}_j \in \mathbb{R}^{d_j \times v_j}$ is a matrix of learnable weights. Essentially, this operation describes a lookup search in a dictionary with $v_j$ entries, where each entry contains a learnable $d_j$-dimensional vector of embeddings that characterises each of the possible values that $q_j$ can take. To obtain a global collection of embeddings $\mathbf{e}$, we concatenate all the vectors $\mathbf{e}_j$ for each categorical covariate $j$:
\begin{equation*}
    \mathbf{e} = \Big\Vert_{j=1}^c \mathbf{e}_j
\end{equation*}
where $c$ is the number of categorical covariates and $\Vert$ represents the concatenation operator. We then use the learnable embeddings $\mathbf{e}$ in downstream tasks.

In terms of the player's architecture, we model both the generator $G$ and discriminator $D$ as neural networks that leverage independent instances $\mathbf{e}^{G}$ and $\mathbf{e}^{D}$ of the categorical embeddings for their corresponding downstream tasks. Specifically, we model the two players as follows:  
\begin{equation*}
        G_{\theta}(\mathbf{z}, \mathbf{r}, \mathbf{q}) = \text{MLP}(\mathbf{z} \Vert \mathbf{r} \Vert \mathbf{e}^{G}) \quad \  D_{\omega}(\bar{\mathbf{x}}, \mathbf{m}, \mathbf{r}, \mathbf{q})  = \text{MLP}(\bar{\mathbf{x}} \Vert \mathbf{m} \Vert \mathbf{r} \Vert \mathbf{e}^{D})
\end{equation*}
where $\text{MLP}$ denotes a multilayer perceptron.

\section{Simulations}
\subsection{Clinical case studies}
The clinical case studies used for the simulations are derived from \cite{barbiero2020computational}. The first scenario consists of an elderly patient experiencing hypertension and type 2 diabetes with diabetic nephropathy. Her lifestyle is mainly sedentary and her diet is rich in carbohydrates. The patient needs a therapeutic plan for the treatment of her hypertension. The task for the clinician is to personalise the therapy according assigning a proper daily dosage of Benazepril. This case study is used to show how the digital patient model can be employed to simulate the evolution over time of clinical endpoints under a set of possible therapeutic plans and to choose the best option. 
In the second scenario the same patient is seeking medical help for a mild flu caused by a SARS-CoV infection. 
For this case study the model can be used to constantly monitor and forecast clinical endpoints to prevent complications threatening patient's life.
The decreased oxygenation caused by flu may have detrimental effects on both heart and brain activities indeed. Studies have reported that SARS-CoV infections can activate the blood clotting pathway by impairing left heart pumping performance which results in a blood back up in the lungs and in a increased blood pressure. High blood pressure can reduce blood vessel's compliance decreasing blood and oxygen flows and leading to a higher risk of developing systemic conditions.
For this reason heparin-based therapies have been recommended to prevent clot formation or tissue plasminogen activator (tPA) \cite{Sardu2020,Tang2020}.
Although some variation in blood pressure throughout the day is normal, a high blood pressure variability is associated with a higher risk of cardiovascular disease \cite{Wen2015,ORourke2005,Mitchell2010,Bangalore2020,Clark2019} and all-cause mortality \cite{Kim2018,Tao2017}.
Clogged arteries, fibrosis, and strokes caused by blood pressure spikes are among the main complications threatening patient's life and calling for the foremost necessity for treatment. 
Hence, blood pressure is one of the most relevant clinical endpoints which need to be constantly monitored in real time and accurately forecast.

\input{results_GNN}

\subsection{Forecasting clinical scenarios and interpreting GNN simulations}
The proposed GNN-based model presented in Section \ref{sec:model} is hereby used to actively monitor and forecast the endpoints highlighted in the two clinical case studies. First, the computational ODE-based system described in \cite{barbiero2020computational} is used to generate a time series for each differential equation with a window size of $\tau=500$ time steps \cite{barbiero2020computationalcode}. Time series are collected, randomly shuffled, and stacked in a dataset. Each item of the collection is randomly assigned either to a training ($n_{train}=3200$), validation ($n_{val}=800$), or test set ($n_{test}=1000$). 

The graph model is derived from the structure of ODE system, thus leveraging human knowledge. 
Nodes correspond to variables represented by the differential equations in \cite{barbiero2020computational} while edges follow the underlying relationships. In a GNN-based model, each node learns a latent representation of the state using the messages received from its neighbourhood. Hence, the rigid mathematical structure of the ODE system is relaxed in our model as such structure can be learned directly from data. The learning process lasts for $\eta=50$ epochs with a learning rate of $\epsilon=0.01$. 
Once trained and validated, the model is used to generate a bundle of possible trajectories for elements of the test set using the procedure described in Section \ref{sec:uncertainty}. As a result, the model estimates a $95\%$ confidence interval of the evolution of each variable over time. 

Providing a complete overview of the clinical state of a patient is not trivial. Arguably, focusing just on one endpoint might be misleading. On the contrary, a global vision comprising pathophysiological conditions is required in order to provide a clear and effective overview where organs and physiological systems can be monitored as a whole. One of the most effective approaches consists in applying a dimensionality reduction technique \cite{van2009dimensionality} condensing the information of each organ and projecting forecasts in a lower-dimensional space. Figure \ref{fig:gnn_results} shows an overview of the clinical state of the heart in a two-dimensional phase space.
For each clinical case study, a GNN-based model is used to simulate a therapeutic intervention and its impact on blood pressure in heart chambers (right and left atrium and ventricle). 
In order to provide an overview of heart conditions, we projected the predicted trajectories using principle component analysis (PCA, \cite{pearson1901lines}). The interpretation of both pictures is straightforward. The first one shows the effect of a therapeutic intervention comprising an increased physical exercise, a reduced amount of calorie intake, and the subscription of a daily dosage of Benazepril (5mg). The predicted result of the prescription (green density reporting the 95\% CI of the trajectories) reveals an overall reduction of blood pressure mean and variability in heart chambers. This results in a reduced risk of developing severe cardiovascular conditions with detrimental ripple effects for the whole system. The second figure, instead, reports the simulation corresponding to the second case study. The same patient is seeking medical help to treat the first symptoms of a SARS-CoV-2 infection. The first simulation (red density) shows the long-term impact on heart blood pressure of an untreated viral infection. In this case, blood pressure spikes may cause irreparable damages to blood vessel walls, reducing their compliance, and impairing their capacity for adaptation to different environmental conditions. A synergic therapy including both Benazepril (5 mg/day) and intra venous injection of heparin (5000 U/ml) may have a beneficial effect on blood pressure mean and variability (orange density). On one side, Benazapril lowers blood pressure by inhibiting ACE activity in cleaving ANG-I and producing ANG-II which is the key RAS regulator of blood pressure. On the other hand, heparin is used to prevent and dissolve blood clots \cite{Sardu2020,Tang2020}. The treatment has an indirect impact on blood pressure by making blood less dense, reducing clotting formation, and lowering inflammation.

A lower-dimensional representation of an organ or system as a whole could be interesting to get a rapid and clear overview of the long-term impact of a disease or a therapeutic intervention. Nonetheless, bundle of predicted trajectories can be visualised and monitored individually in real time when needed in order to investigate patterns in the time domain. Figure \ref{fig:gnn_results} shows an example where blood pressure trajectories in heart chambers are predicted in real time starting from a healthy state condition (green density).
In some cases, this representation in the time domain might be closer to common clinical approaches, thus providing a more conventional visualisation tool for monitoring clinical endpoints in real-time.

    
    
    
    

\subsection{Transcriptomics analysis of the crosstalk between tissue-types}\label{sec:applications_crosstalk}

We hypothesise that the communication factors in blood might be playing an important role in the development of the SARS-CoV-2 infection by facilitating the spread of the virus in the human body. Here, we study whether the expression of genes involved in the renin-angiotensin system can be explained by genes that take part of the communicome in blood. This analysis might shed light on whether it is sensible to model the crosstalk between tissue types with a GNN where tissue nodes communicate with each other through whole blood.


\textbf{Dataset. } We leverage data from the Genotype-Tissue Expression (GTEx) project (v8), a resource that has generated a comprehensive collection of human transcriptome data in a diverse set of tissues \cite{aguet2019gtex}. The dataset contains 15,201 RNA-Seq samples collected from 49 tissues of 838 unique donors. We select genes based on expression thresholds of $\ge 0.1$ TPM in $\ge 20$\% of samples and $\ge 6$ reads in $\ge 20$\% of samples. We normalise the read counts between samples using the trimmed mean of M-values (TMM) normalisation method \cite{robinson2010scaling} and we inverse normal transform the expression values for each gene. From all the donors, we select those that have gene expression measurements for whole blood, yielding 670 unique individuals. We then match the patients' whole blood samples with the corresponding measurements in lung (418), cortex of kidney (62), pancreas (257), and left ventricle of heart (324). Finally, we use the KEGG pathway database \cite{kanehisa2010kegg} to select genes from the renin-angiotensin system (\emph{hsa04614}), chemokine (\emph{hsa04062}), TNF (\emph{hsa04668}), and TGF-$\beta$ (\emph{hsa04350}) pathways.  




\textbf{Model. } We model the expression of genes from the renin-angiotensin system in lung, kidney, pancreas, and heart as a function of genes in the chemokine, TNF, and TGF-$\beta$ pathways in blood. Let $\mathbf{Y} = (Y_1, ..., Y_n)^{\top}$ and $\mathbf{X} = (X_1, ..., X_m)^{\top}$ be multivariate random variables representing the expression of the $n$ genes in the renin-angiotensin system and the $m$ genes in the signalling pathways, respectively. Our model is based on ridge regression \cite{hoerl1970ridge}:
\begin{equation*}
    \mathbf{Y} = \mathbf{X}\mathbf{W} + \bm{\epsilon}
\end{equation*}

where $\mathbf{W} \in \mathbb{R}^{m \times n}$ is a matrix of learnable weights and $\bm{\epsilon} \in \mathbb{R}^{n}$ are the residuals. We optimise the following objective:
\begin{equation*}
    \min_{\mathbf{W}} ||\mathbf{Y} - \mathbf{X}\mathbf{W}||^2_2 + \alpha ||\mathbf{W}||^2_2
\end{equation*}

where $\alpha$ is a hyperparameter that controls the regularisation strength. We also tried non-linear models such as support vector machines, Gaussian processes, and random forests, but they were not significantly better than ridge regresssion according to our cross-validation scores.


\textbf{Results. } Figure \ref{fig:association_results} show the bootstrapped $R^2$ scores for each gene in the renin-angiotensin system pathway in different tissue types. These results show that the expression of some genes in the ACE2 pathway can be partially explained by signalling genes from whole blood. Notably, the associations for the kidney (cortex) are weaker or nonexistent, potentially because the data is limited for this tissue (62 samples) or because the biological associations are indeed small. Overall, these results suggest that signalling pathways such as TNF, TGF-$\beta$, and chemokine might be playing an important role in the development of the SARS-CoV-2 infection.

\input{results_gene_associations}

\subsection{Generative model for transcriptomics data}

The generative model developed in Section \ref{sec:gan_generative} is here used to produce synthetic transcriptomics data. By conditioning on high expression of ACE2 in the lung, kidney, and pancreas, we aim to simulate the effects of SARS-CoV-2 infection in the expression of genes involved in communicome and signalling pathways such as TNF, TGF-$\beta$ and chemokines. These pathways are implicated in many physiological and pathological processes including the regulation of blood pressure and inflammatory processes, and has been hypothesised to play a central role in SARS-CoV-2 infection \cite{Xiao2018,garvin2020mechanistic}.

\textbf{Dataset. } We use data from the GTEx project and process it as described in Section \ref{sec:applications_crosstalk}. 

\begin{figure}
    \centering
    
    \includegraphics[width=0.49\columnwidth]{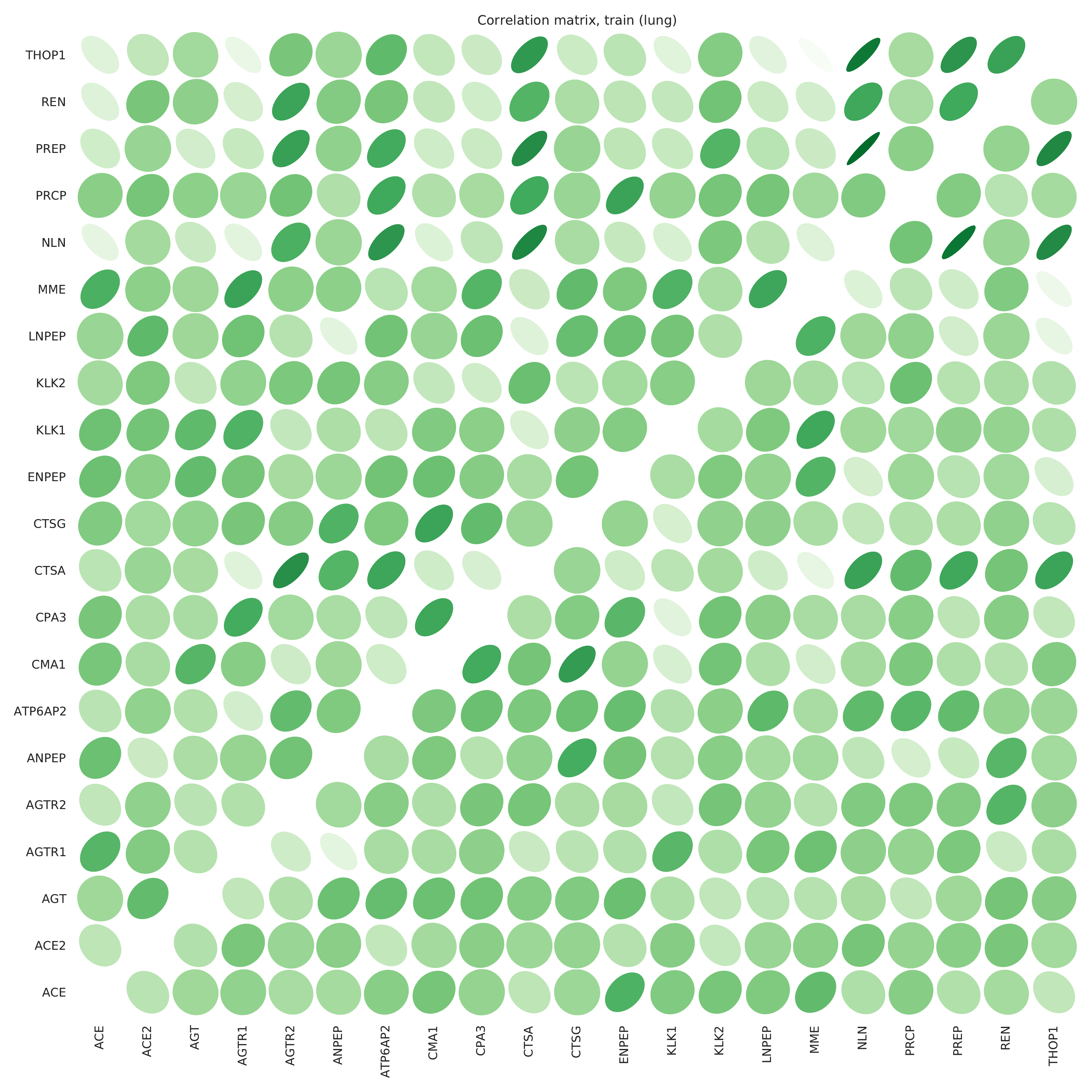}
    \includegraphics[width=0.49\columnwidth]{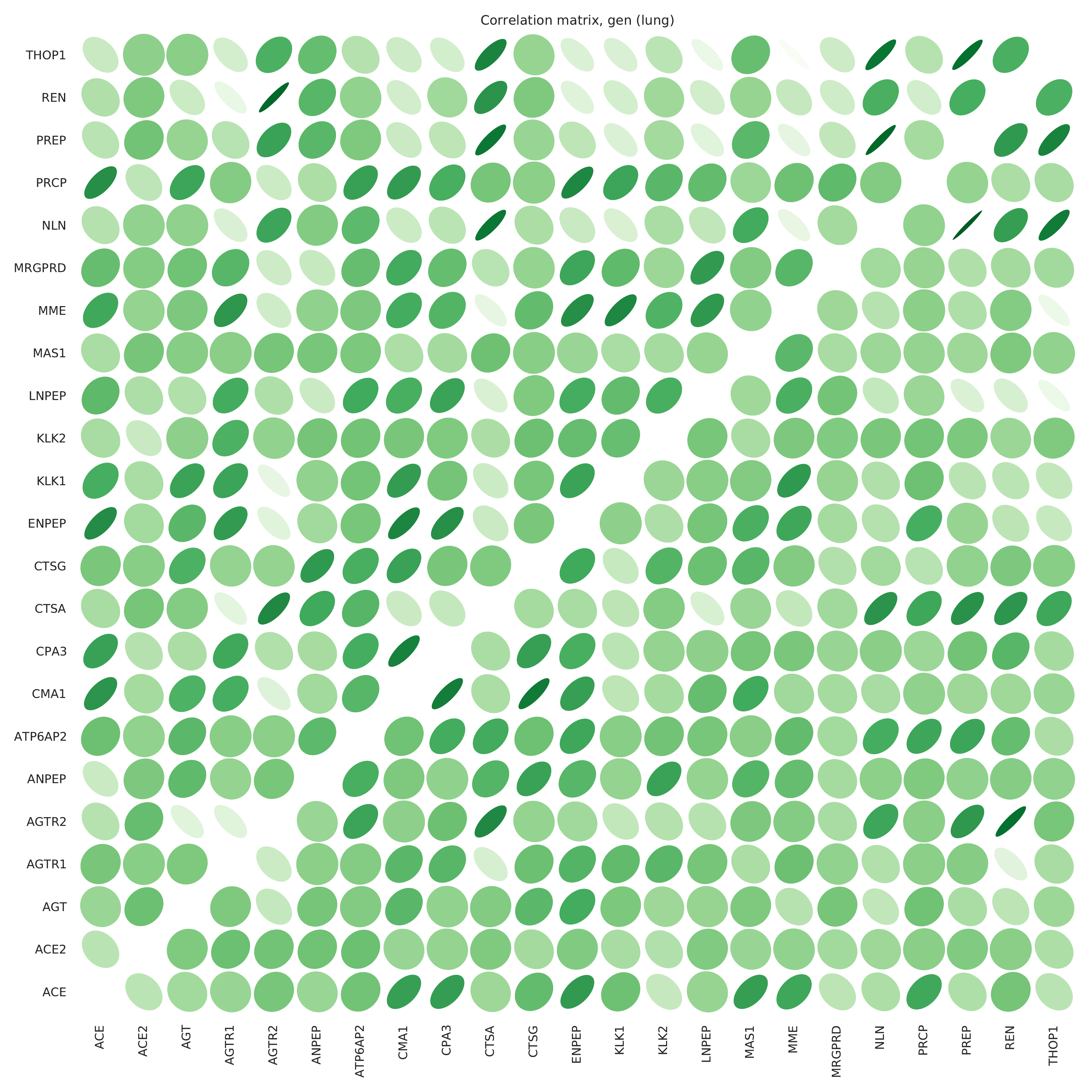}
    
    \caption{Pairwise correlations between genes in the renin-angiotensin system pathway in lung for real (left) and synthetic (right) data. The correlations in the lower and upper matrices are computed from samples with low (60 samples) and high (61 samples) ACE2 expression, respectively. }
    \label{fig:corr_lung_ace2_high_low}
\end{figure}

\begin{figure}
    \centering
    
    \includegraphics[scale=0.5]{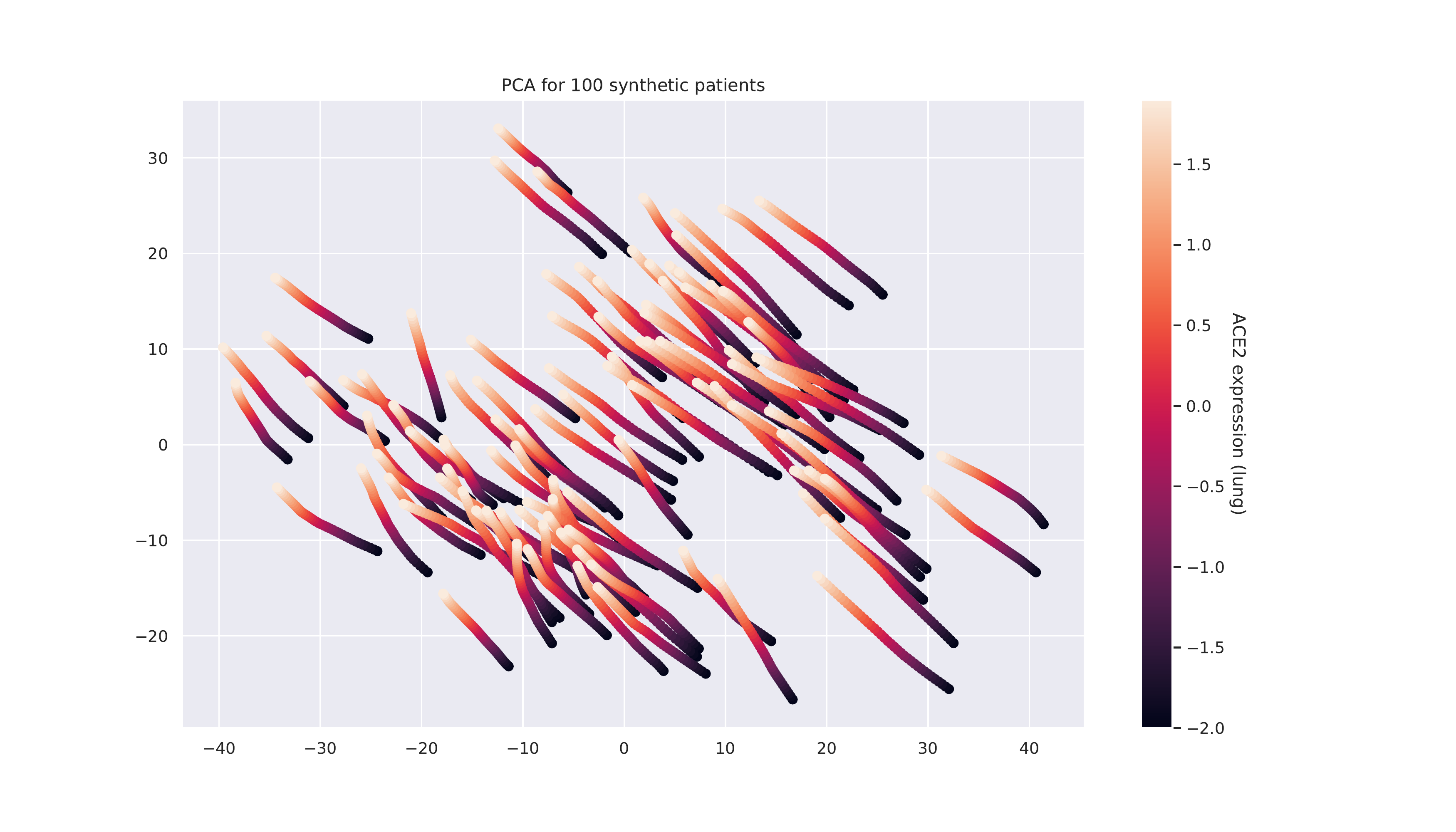}
    
    \caption{Principal component analysis of the multi-tissue expression of 100 synthetic patients for different levels of ACE2 expression. Each line corresponds to a unique patient. For each patient, we fix all the latent covariates and modify the levels of ACE2 in lung. Overexpressing ACE2 leads to changes in the expression of other genes and these changes follow a well-defined trajectory.}
    \label{fig:umap_ace2_lung}
\end{figure}

\textbf{Results. } Figure \ref{fig:corr_lung_ace2_high_low} shows that the pairwise correlations between genes in the ACE2 pathway (lung) are well preserved in the synthetic data. We observe that some genes in the renin-angiotensin system pathway (CTSA, AGTR2, NLN, and PREP) that can be relatively well explained as a function of blood signalling factors (see Figure \ref{fig:association_results}) are simultaneously correlated with ACE2. This suggests that these genes might be playing an important role in the spread of SARS-CoV-2 in our body through blood. Figure \ref{fig:umap_ace2_lung} shows that it is possible to sample data for synthetic patients conditioned on different levels of ACE2 expression in lung. 

\subsection{Software}
All the code for the experiments has been implemented in Python 3, relying upon open-source libraries \cite{abadi2016tensorflow,pedregosa2011scikit,wang2019deep}.
All the experiments have been run on the same machine: Intel\textsuperscript{\textregistered} Core\texttrademark\ i7-8750H 6-Core Processor at 2.20 GHz equipped with 8 GiB RAM.
To enable code reuse, the Python code for the mathematical models including parameter values and documentation is freely available under GNU Public License from a GitHub repository\footnote{\url{https://github.com/pietrobarbiero/digital-patient}} \cite{barbiero2020digitalcode}. 
Unless required by applicable law or agreed to in writing, software is distributed on an "as is" basis, without
warranties or conditions of any kind, either express or implied.

\section{Discussion and conclusion}

\subsection{Interplay between GAN and GNN models}
The models presented in this work (GAN, GNN) are independent of each other. On one hand, the main goal of the GNN model (see Section \ref{sec:graph_nn}) is to forecast various patient's conditions based on real or synthetic data, integrating information that spans multiple layers of the human body. On the other hand, the GAN model (see Section \ref{sec:gan_generative}) is able to generate data under different states, effectively enriching the space of pathophysiological conditions and endowing the digital twin with the ability to simulate the effects of counterfactual events. The independence of these two models enables a modular framework wherein each module can be trained separately on a distinct data modality. Importantly, these modules can be composed and reused through transfer learning. In this work, we have shown how computational models can be used to generate synthetic training data representing physiological conditions. Following the same principles, each module of a complex architecture could be pre-trained on synthetic simulations, refined using data obtained from horizontal population studies, and finally personalised according to clinical health records.

The GAN and the GNN models can be interconnected in a synergistic way. In order to train the GNN effectively, it is necessary to have access to  heterogenous, paired data modalities (from different layers: genomic, transcriptomic, cellular, organ, exposomic, ...) collected from a comprehensive collection of patients and encompassing a wide variety of conditions. However, to the best of our knowledge, to this date no such dataset exists. This is mainly because collecting paired, multilayer data from patients is expensive and entails important ethical and privacy concerns \cite{jobin2019global,mittelstadt2019principles}. To address this issue, our GAN framework can synthesise data at multiple layers conditioned on the patient's conditions (e.g. diabetic, hypertension, ...) and clinical information (e.g. heart rate, blood pressure, age, sex, ethnicity, exercise, nutrition, ...). This synthetic data can be used to train the GNN and impute missing data modalities of real patients.


\subsection{Advantages, limitations and visions}


\emph{The promise of artificial intelligence in medicine is to provide composite, panoramic views of individuals' medical data; to improve decision making; to avoid errors such as misdiagnosis and unnecessary procedures; to help in the ordering and interpretation of appropriate tests; and to recommend treatment} \cite{topol2019deep}. The future of medicine is already bound to AI. Technological innovations are completely changing medicine perspectives expanding its horizons and moving towards an holistic view of human beings. The destiny of the whole healthcare system depends on this radical paradigm shift. Embracing AI innovations is just a technological prerequisite, the first step towards a total transformation of how medicine currently works, is delivered, and perceived by patients. Thinking that AI will just and mainly improve clinical decision making is wrong. AI may actually open the doors to completely new ways of investigating the human body as a whole. The core and ultimate purpose of health will be developing preventative and personalised pathways to well-being rather than delivering treatments. This does not mean that medicine will no longer be connected to illness. On the contrary, the future foreseen is that AI will assist medicine in improving diagnosis and devising novel therapeutic strategies to deliver more effective solutions. The current healthcare revolution will not take back all the past technological advances, but it will show them under a new light. 

Ethical repercussions will also be huge \cite{jobin2019global,mittelstadt2019principles}. The transition has begun, but it will call for deeper trans-disciplinary research and a substantial technological innovation in a variety of research areas. Education will also play a key role in changing lifestyle habits and the way health is perceived, communicated, and delivered \cite{yu2017towards}. From a clinical standpoint, AI will support a plethora of different tasks from medical check up to personalised intervention strategies to contrast ripple effects or to promote healthy habits. In non acute states, predictive inference will propose prevention plans for comorbidity management, particularly in presence of multiple therapies \cite{rivera2020guidelines}. 

Increasingly large amount of personal data will be collected to feed modular machine learning (ML) models organised to address specific and personalised medical issues. Clinical endpoints will be constantly monitored, shared, and compared in order to answer relevant research questions and to deliver the best possible service. A deeper understanding and practice of modelling in medicine
will produce better investigation of complex biological processes, and even new ideas and better feedback into medicine. Modelling-based approaches combined with data-driven ML techniques will progressively provide models with higher degree of interpretability and generalisation ability \cite{barbiero2020modeling} which will make evidence-based medicine even more accessible intensifying the involvement of patients in the decision making process. 
Besides, for each individual both healthcare systems and private companies will collect, save, and eventually exploit an enormous amount of personal data. Providing an effective, stable, and unified juridical overview is critical on this matter \cite{panch2019inconvenient}.

AI simulations forecasting the evolution of clinical endpoints over time will also reshape clinical guidelines \cite{rivera2020guidelines} which will no longer be based just on \emph{horizontal} population studies. Cross-modality data will be collected for each patient and machine learning models will be used to predict a bundle of possible trajectories representing the future states of the patient allowing for personalised prescriptions, surgical planning and medical interventions.
Finally, AI will change the leading \emph{vehicle} of medicine. The demand for AI-powered and IoT devices is increasing worldwide. The future medical equipment will likely required to be cheap and extremely modular, but more importantly it needs to be deployable in dedicated hardware to be distributed in larger markets.

    The most important question around the integration of AI in medicine is about the benefit for the patient. A meaningful quote about twins is the following: being a twin is like being born with a best friend. The data integration will make a better portrait of patient's condition trajectories but will require data inter-operability and data security. Technology is often not neutral, but transformed to be biased in one way or another \cite{ellul1954technique}. Individuals can have different unforeseen readings and usage of new technologies. It may increase both user vulnerability and user empowerment. The vulnerability is the combination of exposure to the variety of personal medical data and the coping capabilities of users which could be different between young and mature people, as young are usually quicker in incorporating a new technology into everyday life. The user is empowered if he/she acquires awareness and control of his/her condition and context. A common example are online (website and blogs) initiatives such as patientslikeme which allow the user to search and make up his/her mind about a disease \cite{wicks2010sharing}. Instead the user disempowerment depends on the lack of technical knowledge of how mechanisms work; this is even enhanced in black box techniques such as deep learning.\\  
    The second question is about impact on how clinicians work or are trained. We believe that improving both data integration and predictability will provide physicians with improved  medical decisions support systems and a decrease in both costs, through the evaluation of best therapies, and errors. A limitations is the poor interpretability and explainability in deep learning architectures. This limitation will also greatly affect the training of the new clinicians on AI technologies. There are growing efforts to make neural networks more interpretable in order to keep the human (doctors and patients) in the loop. The interpretability could be improved by using in parallel mechanistic computational modelling and simulations (\cite{Bartocci2016,Milanesi2009}), model extraction libraries (see for instance \cite{kazhdan2020marleme}), and visual inference tools \cite{bodnar2020deep}.
    This tool could also be complemented by clinical decision support systems such as \cite{Mller2020}.\\
    The complexly structured and multi level comorbidity and frailty patterns of most diseases describe a highly dynamical system and are, therefore, challenging current medical therapies.

Mechanistic computational modelling and machine learning should be considered together when building innovative healthcare solutions. Building a puzzle is often an example of participatory activity. Clinicians, mechanistic computational modelling and machine learning researchers, data policy makers, public and private sectors could build a puzzle (i.e. the healthcare) together and they should first develop a shared vision about what is the puzzle. 
Our vision is to consider a co-simulation (say doctor checkup visits vs computational experiments) of the two twins to allow co-verification. From a theoretical computer science perspective, this could open the direction of an interplay between AI and  verification/synthesis and the use of reachability analysis to identify constraints over the well-being and disease system state space.
Although different architectures seem suitable (e.g. only GNNs, only GANs, VAEs, etc), our design has important advantages: the GNN could provide a physical mapping of the human body (in the same way a tube map or bus route is a map of a city); GANs could be specialised on processing molecular information or they could operate cross modal operations such as omic-omic, omic-clinical, clinical-clinical.


In this work we demonstrate the feasibility of representing and integrating physiological models and molecular information using graph neural networks and generative adversarial networks. This composite approach provides modularity and scalability across layers of biomedical data, it is amenable of a battery of modeling approaches, and generates integrated predictions which translate into patients trajectories. We have assimilated our product to a digital twin of the patient.


\section{Acknowledgement}
This work has received funding from the European Union’s Horizon 2020 research and innovation programme under grant agreement No 848077. This project has received funding from "la Caixa" Foundation (ID 100010434), under agreement LCF/BQ/EU19/11710059.

\bibliographystyle{unsrt}  
\bibliography{references}

\end{document}

%% file: architecture.tex
\begin{figure}
\centering
\newcommand\initialy{4}
\newcommand\nodeSize{0.75cm}

\tikzset{%
  tipSquare/.tip={Circle[open]}
}

\tikzset{%
  every neuron/.style={
    circle,
    draw,
    fill=white,
    scale=0.8,
    minimum size=\nodeSize
  },
  neuron missing/.style={
    draw=none, 
    scale=0.8,
    fill=$\dots$,
    text height=0cm,
    execute at begin node=$\dots$
  },
  snake it/.style={
    decorate, decoration=snake
  }
}

\begin{tikzpicture}[x=1.5cm, y=1.5cm, >=stealth, scale=0.54, every node/.style={transform shape}, curved arrow/.style={arc arrow={to pos #1 with length 2mm and options {}}},
reversed curved arrow/.style={arc arrow={to pos #1 with length 2mm and options reversed}}]  

\def\ystart{-1.3}
\def\xstart{-2.8}
\draw[rounded corners, densely dotted, fill=blue!1] (\xstart,\ystart) -- (\xstart+11.8,\ystart) -- (\xstart+11.8, \ystart-16.8) -- (\xstart, \ystart-16.8) -- cycle;

\def\ystart{4.6}
\def\xstart{-1.9}
\draw[rounded corners, densely dotted, fill=blue!1] (\xstart,\ystart) -- (\xstart+3.5,\ystart) -- (\xstart+3.5, \ystart-5.2) -- (\xstart, \ystart-5.2) -- cycle;
\def\ystart{4.4}
\def\xstart{-1.7}
\draw[rounded corners, densely dotted, thick, red, fill=blue!1] (\xstart,\ystart) -- (\xstart+3.1,\ystart) -- (\xstart+3.1, \ystart-2) -- (\xstart, \ystart-2) -- cycle;
\def\ystart{2.2}
\def\xstart{-1.7}
\draw[rounded corners, densely dotted, thick, blue, fill=blue!1] (\xstart,\ystart) -- (\xstart+3.1,\ystart) -- (\xstart+3.1, \ystart-1.6) -- (\xstart, \ystart-1.6) -- cycle;
\def\ystart{0.4}
\def\xstart{-1.7}
\draw[rounded corners, densely dotted, thick, black!60!green, fill=blue!1] (\xstart,\ystart) -- (\xstart+3.1,\ystart) -- (\xstart+3.1, \ystart-0.8) -- (\xstart, \ystart-0.8) -- cycle;

\def\ystart{-14.1}
\def\xstart{-6.2}
\draw[rounded corners, densely dotted, fill=blue!1] (\xstart,\ystart) -- (\xstart+3.3,\ystart) -- (\xstart+3.3, \ystart-0.8) -- (\xstart, \ystart-0.8) -- cycle;
\node[align=center, above] at (-4.5, -14.7) {$\mathbf{y}^{(n)} = \mathbf{F}(t, \mathbf{y}, \mathbf{y}', \dots, \mathbf{y}^{(n-1)})$};
\node[align=center, above] at (-4.5, -15.3) {ODE system};

\node[inner sep=0pt] (chip) at (3,0.2)
    {\includegraphics[width=.15\textwidth]{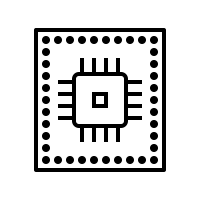}};
\node[inner sep=0pt] (data) at (3,2.8)
    {\includegraphics[width=.15\textwidth]{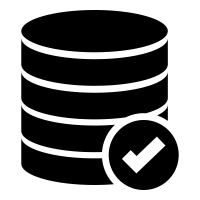}};
\node[inner sep=0pt] (data) at (-3.4,1.8)
    {\includegraphics[width=.22\textwidth]{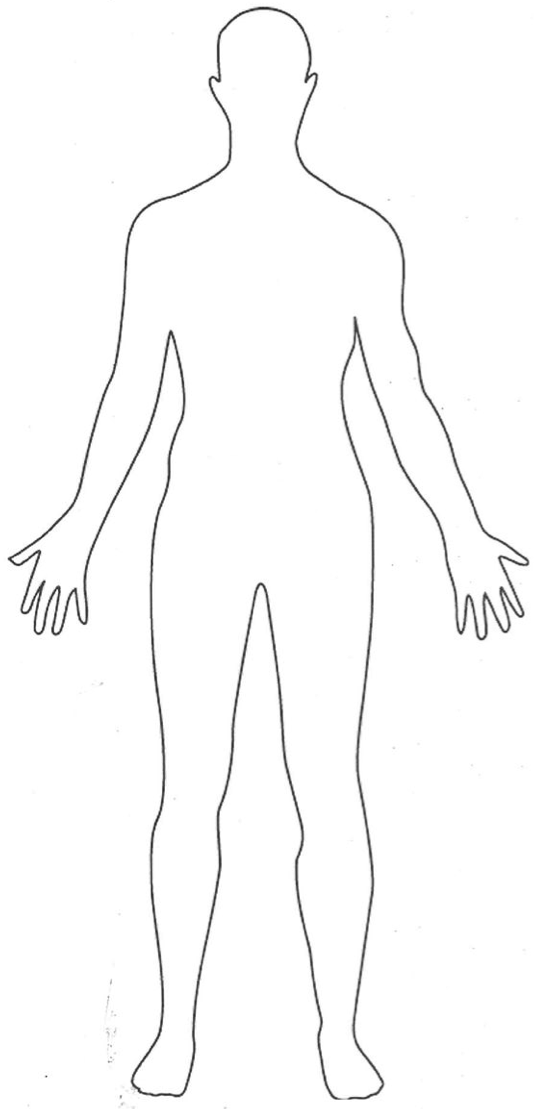}};
\node[align=center, above] at (-3.4,-1.2) {patient};
\node[inner sep=0pt] (data) at (1,3.8)
    {\includegraphics[width=.07\textwidth]{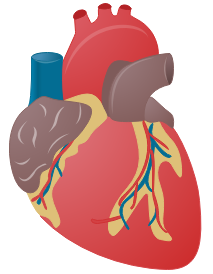}};
\node[inner sep=0pt] (data) at (-1,3.5)
    {\includegraphics[width=.07\textwidth]{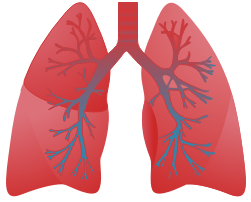}};
\node[inner sep=0pt] (data) at (0,3)
    {\includegraphics[width=.07\textwidth]{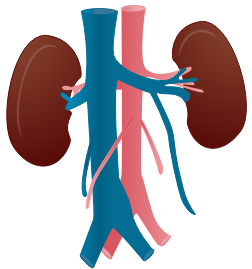}};
\node[inner sep=0pt] (data) at (-0.8,1.6)
    {\includegraphics[width=.07\textwidth]{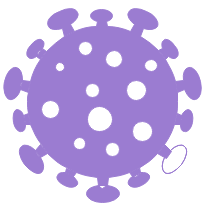}};
\node[inner sep=0pt] (data) at (0.8,1.1)
    {\includegraphics[width=.07\textwidth]{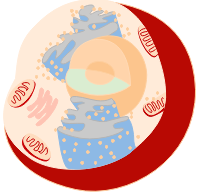}};
\node[inner sep=0pt, rotate=90] (data) at (0,0)
    {\includegraphics[width=.04\textwidth]{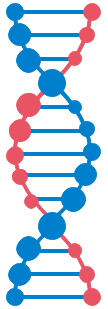}};

\def\xstart{4.9}
\def\ystart{0.2}
\draw[->] (\xstart-0.2,\ystart+0) -- (\xstart+3.2,\ystart+0) node[right] {$t$}; 
\draw[domain=\xstart:\xstart+3, smooth, variable=\x, color=black!50!green!90] plot (\x,{0.3*sin(5 * \x r) + \ystart}) node[right] {healthy trajectory};
\draw[domain=\xstart:\xstart+3, smooth, variable=\x, color=black!50!red!90] plot (\x,{0.8*sin(5 * \x r) + \ystart}) node[right] {disease};

\def\xstart{3}
\draw[rounded corners] [->] (\xstart, -1.3) -- (\xstart, -0.5);
\def\xstart{3}
\draw[rounded corners] [<->] (\xstart, 2) -- (\xstart, 0.8);
\def\ystart{0.2}
\draw[rounded corners] [->] (3.5, \ystart) -- (4.5, \ystart);
\def\ystart{0.2}
\draw[rounded corners] [->] (1.6, \ystart) -- (2.4, \ystart);
\def\ystart{2.8}
\draw[rounded corners] [->] (1.6, \ystart) -- (2.2, \ystart);
\draw[-{Circle[]}, black!50!red, dotted] (0.6,4) -- (-2.3,4) -- (-3.4,3.1);
\draw[black!50!red, dotted] [<-] (0.6,4) -- (-2.3,4) -- (-3.4,3.1);
\draw[-{Circle[]}, black!50!red, dotted] (-1.4,3.3) -- (-3.5,2.8);
\draw[black!50!red, dotted] [<-] (-1.4,3.3) -- (-3.5,2.8);
\draw[-{Circle[]}, black!50!red, dotted] (-0.4,2.9) -- (-3.4,2.5);
\draw[black!50!red, dotted] [<-] (-0.4,2.9) -- (-3.4,2.5);
\draw[-{Circle[]}, black!50!blue, dotted] (-1.3,1.7) -- (-3.7,2.3);
\draw[black!50!blue, dotted] [<-] (-1.3,1.7) -- (-3.7,2.3);
\draw[-{Circle[]}, black!50!blue, dotted] (0.3,1) -- (-2.6,1) -- (-3.3,1.9);
\draw[black!50!blue, dotted] [<-] (0.3,1) -- (-2.6,1) -- (-3.3,1.9);
\draw[-{Circle[]}, black!50!blue, dotted] (0.3,1) -- (-2.6,1) -- (-3.3,2.4);
\draw[black!50!blue, dotted] [<-] (0.3,1) -- (-2.6,1) -- (-3.3,2.4);
\draw[-{Circle[]}, black!50!green, dotted] (-0.8,0) -- (-2.8,0) -- (-3.6,2.2);
\draw[black!50!green, dotted] [<-] (-0.8,0) -- (-2.8,0) -- (-3.6,2.2);
\draw[-{Circle[]}, black!50!green, dotted] (-0.8,0) -- (-2.8,0) -- (-4.4,1.8);
\draw[black!50!green, dotted] [<-]  (-0.8,0) -- (-2.8,0) -- (-4.4,1.8);
\draw[rounded corners] [->] (-3.9,-1) -- (-4.5,-1) -- (-4.5,-14);
\draw[rounded corners] [->] (-4.5,-15.3) -- (-4.5,-16.4) -- (-2.55, -16.4);

\def\ystart{-2.5}
\draw[rounded corners,fill=gray!6] (1.7,\ystart) -- (7.4,\ystart) -- (7.4, \ystart-4) -- (1.7, \ystart-4) -- cycle;

\def\ystart{-9.5}
\draw[rounded corners,fill=gray!6] (0.5,\ystart) -- (6.2,\ystart) -- (6.2, \ystart-4) -- (0.5, \ystart-4) -- cycle;

\def\ystart{-15}
\draw[rounded corners,fill=gray!6] (1,\ystart+0.5) -- (5.9,\ystart+0.5) -- (5.9, \ystart-2.5) -- (1, \ystart-2.5) -- cycle;
\draw[rounded corners] [->] (0.5, -17) -- (1.5, -17);
\node[every neuron](phie) at (1.7, -17){$\phi^e$};
\draw[rounded corners] [->] (0.5, -16) -- (3.3, -16);
\draw[rounded corners] [->] (0.5, -16) -- (1.5, -16.9);
\node[every neuron](phiv) at (3.5, -16){$\phi^h$};
\draw[rounded corners] [->] (1.9, -16.9) -- (3.3, -16.1);
\def\ystart{-16.2}
\draw[rounded corners,fill=white] (2.3,\ystart) -- (3,\ystart) -- (3, \ystart-0.5) -- (2.3, \ystart-0.5) -- cycle;
\node[align=center, above] at (2.65, -16.65) {$\rho^{e\rightarrow h}$};
\node[every neuron](phiu) at (5.3, -15){$\phi^e$};
\draw[rounded corners] [->] (3.7, -15.9) -- (5.1, -15.1);
\def\ystart{-15.2}
\draw[rounded corners,fill=white] (4.1,\ystart) -- (4.8,\ystart) -- (4.8, \ystart-0.5) -- (4.1, \ystart-0.5) -- cycle;
\node[align=center, above] at (4.45, -15.65) {$\rho^{h\rightarrow u}$};
\draw[rounded corners] [->] (3.7, -16) -- (6.5, -16);
\draw[rounded corners] [->] (5.5, -15) -- (6.5, -15);
\draw[dotted] (2.1, -14.5) -- (2.1, -17.5);
\draw[dotted] (3.9, -14.5) -- (3.9, -17.5);
\def\xstart{-2.05}
\def\ystart{-16.2}
\draw[rounded corners,dashed,fill=black!5!green!5] (\xstart,\ystart-1.2) -- (\xstart+2.55,\ystart-1.2) -- (\xstart+2.55, \ystart+0.6) -- (\xstart, \ystart+0.6) -- cycle;
\node[align=center, above] at (-2.4, -16.6) {$\mathbf{G}$};
\def\xstart{-1.5}
\def\ystart{-16.3}
\draw[rounded corners,dashed,fill=black!30!green!20] (\xstart,\ystart) -- (\xstart+1.9,\ystart) -- (\xstart+1.9, \ystart+0.6) -- (\xstart, \ystart+0.6) -- cycle;
\def\xstart{-1.2}
\def\ystart{-16}
\node[every neuron](v11) at (\xstart, \ystart){$h_1$};
\node[every neuron](v12) at (\xstart+0.5, \ystart){$h_2$};
\node[](vv) at (\xstart+0.9, \ystart){$\dots$};
\node[every neuron](v1n) at (\xstart+1.3, \ystart){$h_n$};
\node[align=center, above] at (-1.8, -16.15) {$\mathbf{H}$};
\def\xstart{-1.5}
\def\ystart{-17.3}
\draw[rounded corners,dashed,fill=black!30!green!20] (\xstart,\ystart) -- (\xstart+1.9,\ystart) -- (\xstart+1.9, \ystart+0.6) -- (\xstart, \ystart+0.6) -- cycle;
\def\xstart{-1.2}
\def\ystart{-17}
\node[every neuron](E11) at (\xstart, \ystart){$E_1$};
\node[every neuron](E12) at (\xstart+0.5, \ystart){$E_2$};
\node[](EE) at (\xstart+0.9, \ystart){$\dots$};
\node[every neuron](E1k) at (\xstart+1.3, \ystart){$E_k$};
\node[align=center, above] at (-1.8, -17.15) {$\mathbf{E}$};
\def\xstart{6.5}
\def\ystart{-16.3}
\draw[rounded corners,dashed,fill=black!30!green!20] (\xstart,\ystart) -- (\xstart+1.9,\ystart) -- (\xstart+1.9, \ystart+0.6) -- (\xstart, \ystart+0.6) -- cycle;
\def\xstart{6.8}
\def\ystart{-16}
\node[every neuron](v11i) at (\xstart, \ystart){$h_1'$};
\node[every neuron](v12i) at (\xstart+0.5, \ystart){$h_2'$};
\node[](vvi) at (\xstart+0.9, \ystart){$\dots$};
\node[every neuron](v1ni) at (\xstart+1.3, \ystart){$h_n'$};
\node[align=center, above] at (8.7, -16.15) {$\mathbf{H'}$};
\def\xstart{6.5}
\def\ystart{-15.3}
\draw[rounded corners,dashed,fill=purple!40] (\xstart,\ystart) -- (\xstart+1.9,\ystart) -- (\xstart+1.9, \ystart+0.6) -- (\xstart, \ystart+0.6) -- cycle;
\def\xstart{6.8}
\def\ystart{-15}
\node[every neuron](u11i) at (\xstart, \ystart){$u_1'$};
\node[every neuron](u12i) at (\xstart+0.5, \ystart){$u_2'$};
\node[](uui) at (\xstart+0.9, \ystart){$\dots$};
\node[every neuron](u1ni) at (\xstart+1.3, \ystart){$u_n'$};
\node[align=center, above] at (8.7, -15.15) {$\mathbf{U'}$};
\draw[rounded corners, densely dotted] [->] (4.1, -8) -- (7, -8) -- (7, -14) -- (-0.8, -14) -- (-0.8, -15.6);

\def\ystart{0}



\def\ystart{-2.3}

\def\xstart{1.8}
\draw[rounded corners,dashed,fill=black!50!green!20] (\xstart,\ystart) -- (\xstart+1.9,\ystart) -- (\xstart+1.9, \ystart+0.6) -- (\xstart, \ystart+0.6) -- cycle;

\def\xstart{6}
\draw[rounded corners,dashed,fill=orange!30] (\xstart,\ystart) -- (\xstart+1.4,\ystart) -- (\xstart+1.4, \ystart+0.6) -- (\xstart, \ystart+0.6) -- cycle;

\def\ystart{-2}
\def\xstart{2.1}
\node[every neuron](z11) at (\xstart, \ystart){$z_1$};
\node[every neuron](z12) at (\xstart+0.5, \ystart){$z_2$};
\node[](zz) at (\xstart+0.9, \ystart){$\dots$};
\node[every neuron](z1n) at (\xstart+1.3, \ystart){$z_u$};

\def\xstart{6.3}
\node[every neuron](q1) at (\xstart, \ystart){$q_1$};
\node[](rr) at (\xstart+0.4, \ystart){$\dots$};
\node[every neuron](qk) at (\xstart+0.8, \ystart){$q_c$};

\def\ystart{-3.3}

\def\xstart{-2.2}
\draw[rounded corners,dashed,fill=orange!30] (\xstart,\ystart) -- (\xstart+1.4,\ystart) -- (\xstart+1.4, \ystart+0.6) -- (\xstart, \ystart+0.6) -- cycle;

\def\xstart{4.1}
\draw[rounded corners,dashed,fill=green!10] (\xstart,\ystart) -- (\xstart+1.4,\ystart) -- (\xstart+1.4, \ystart+0.6) -- (\xstart, \ystart+0.6) -- cycle;

\def\ystart{-3}

\def\xstart{-1.9}
\node[every neuron](r1) at (\xstart, \ystart){$r_1$};
\node[](rr) at (\xstart+0.4, \ystart){$\dots$};
\node[every neuron](rk) at (\xstart+0.8, \ystart){$r_k$};

\def\xstart{2.75}
\node[draw, align=center, fill=white](concat) at (\xstart, \ystart){Concat \\ $\big\Vert$};
\draw [<-] (concat) -- ++(0, 0.7);
\draw [<-] (concat.180) -- ++(-3.1,0);
\draw [<-] (concat.0) -- ++(0.9,0);
\draw [->] (concat) -- ++(0,-0.7);

\def\xstart{4.4}
\node[every neuron](r1) at (\xstart, \ystart){$\mathbf{e}_1$};
\node[](rr) at (\xstart+0.4, \ystart){$\dots$};
\node[every neuron](rk) at (\xstart+0.8, \ystart){$\mathbf{e}_c$};

\def\xstart{6.7}
\node[draw, align=center, fill=white](embed) at (\xstart, \ystart){Embed \\ $\mathbf{W}_j^{G} \bar{\mathbf{q}}_j$};
\draw [<-] (embed) -- ++(0, 0.7);
\draw [->] (embed) -- ++(-1.2, 0);

\def\ystart{-4.3}

\def\xstart{1.8}
\draw[rounded corners,dashed,fill=gray!10] (\xstart,\ystart) -- (\xstart+1.9,\ystart) -- (\xstart+1.9, \ystart+0.6) -- (\xstart, \ystart+0.6) -- cycle;

\def\ystart{-4}

\def\xstart{2.1}
\node[every neuron](h11) at (\xstart, \ystart){};
\node[every neuron](h12) at (\xstart+0.5, \ystart){};
\node[](mm) at (\xstart+0.9, \ystart){$\dots$};
\node[every neuron](h1n) at (\xstart+1.3, \ystart){};

\def\ystart{-5.3}

\def\xstart{1.8}
\draw[rounded corners,dashed,fill=gray!20] (\xstart,\ystart) -- (\xstart+1.9,\ystart) -- (\xstart+1.9, \ystart+0.6) -- (\xstart, \ystart+0.6) -- cycle;

\def\ystart{-5}
\def\xstart{2.1}
\node[every neuron](h21) at (\xstart, \ystart){};
\node[every neuron](h22) at (\xstart+0.5, \ystart){};
\node[](mm) at (\xstart+0.9, \ystart){$\dots$};
\node[every neuron](h2n) at (\xstart+1.3, \ystart){};

\def\ystart{-6.3}

\def\xstart{1.8}
\draw[rounded corners,dashed,fill=gray!30] (\xstart,\ystart) -- (\xstart+1.9,\ystart) -- (\xstart+1.9, \ystart+0.6) -- (\xstart, \ystart+0.6) -- cycle;

\def\ystart{-6}

\def\xstart{0}

\def\xstart{2.1}
\node[every neuron](h31) at (\xstart, \ystart){};
\node[every neuron](h32) at (\xstart+0.5, \ystart){};
\node[](mm) at (\xstart+0.9, \ystart){$\dots$};
\node[every neuron](h3n) at (\xstart+1.3, \ystart){};

\def\ystart{-7.3}
\def\xstart{-0.6}
\draw[rounded corners,dashed,fill=pink!50] (\xstart,\ystart) -- (\xstart+1.9,\ystart) -- (\xstart+1.9, \ystart+0.6) -- (\xstart, \ystart+0.6) -- cycle;

\def\ystart{-7}
\def\xstart{-0.3}
\node[every neuron](m1) at (\xstart, \ystart){$m_1$};
\node[every neuron](m2) at (\xstart+0.5, \ystart){$m_2$};
\node[](mm) at (\xstart+0.9, \ystart){$\dots$};
\node[every neuron](mn) at (\xstart+1.3, \ystart){$m_n$};

\def\xstart{2.75}
\node[draw, align=center, fill=white](concat) at (\xstart, \ystart){Mask \\ $\mathbf{m} \odot G$};
\draw [<-] (concat) -- ++(0, 0.7);
\draw [<-] (concat.180) -- ++(-1,0);
\draw [->] (concat) -- ++(0,-0.7);

\def\ystart{-8}

\def\ystart{-8.3}
\def\xstart{-0.6}
\draw[rounded corners,dashed,fill=blue!10] (\xstart,\ystart) -- (\xstart+1.9,\ystart) -- (\xstart+1.9, \ystart+0.6) -- (\xstart, \ystart+0.6) -- cycle;

\def\xstart{1.8}
\draw[rounded corners,dashed,fill=red!30] (\xstart,\ystart) -- (\xstart+1.9,\ystart) -- (\xstart+1.9, \ystart+0.6) -- (\xstart, \ystart+0.6) -- cycle;

\def\ystart{-8}
\def\xstart{-0.3}
\node[every neuron](z11) at (\xstart, \ystart){$x_1$};
\node[every neuron](z12) at (\xstart+0.5, \ystart){$x_2$};
\node[](zz) at (\xstart+0.9, \ystart){$\dots$};
\node[every neuron](z1n) at (\xstart+1.3, \ystart){$x_n$};

\def\xstart{2.1}
\node[every neuron](z11) at (\xstart, \ystart){$\hat{x}_1$};
\node[every neuron](z12) at (\xstart+0.5, \ystart){$\hat{x}_2$};
\node[](zz) at (\xstart+0.9, \ystart){$\dots$};
\node[every neuron](z1n) at (\xstart+1.3, \ystart){$\hat{x}_n$};

\def\ystart{-9}

\def\ystart{-10.3}

\def\xstart{2.9}
\draw[rounded corners,dashed,fill=green!10] (\xstart,\ystart) -- (\xstart+1.4,\ystart) -- (\xstart+1.4, \ystart+0.6) -- (\xstart, \ystart+0.6) -- cycle;

\def\ystart{-10}

\def\xstart{1.55}
\node[draw, align=center, fill=white](concat2) at (\xstart, \ystart){Concat \\ $\big\Vert$};
\draw [<-] (concat2.0) -- ++(0.9,0);
\draw [->] (concat2) -- ++(0,-0.7);

\def\xstart{3.2}
\node[every neuron](r1) at (\xstart, \ystart){$\mathbf{e}_1$};
\node[](rr) at (\xstart+0.4, \ystart){$\dots$};
\node[every neuron](rk) at (\xstart+0.8, \ystart){$\mathbf{e}_c$};

\def\xstart{5.5}
\node[draw, align=center, fill=white](embed2) at (\xstart, \ystart){Embed \\  $\mathbf{W}_j^{D} \bar{\mathbf{q}}_j$};
\draw [->] (embed2) -- ++(-1.2, 0);

\def\ystart{-11.3}

\def\xstart{0.6}
\draw[rounded corners,dashed,fill=gray!10] (\xstart,\ystart) -- (\xstart+1.9,\ystart) -- (\xstart+1.9, \ystart+0.6) -- (\xstart, \ystart+0.6) -- cycle;

\def\ystart{-11}

\def\xstart{0.9}
\node[every neuron](hd11) at (\xstart, \ystart){};
\node[every neuron](hd12) at (\xstart+0.5, \ystart){};
\node[](mm) at (\xstart+0.9, \ystart){$\dots$};
\node[every neuron](hd1n) at (\xstart+1.3, \ystart){};

\def\ystart{-12.3}

\def\xstart{0.6}
\draw[rounded corners,dashed,fill=gray!20] (\xstart,\ystart) -- (\xstart+1.9,\ystart) -- (\xstart+1.9, \ystart+0.6) -- (\xstart, \ystart+0.6) -- cycle;

\def\ystart{-12}

\def\xstart{0.9}
\node[every neuron](hd21) at (\xstart, \ystart){};
\node[every neuron](hd22) at (\xstart+0.5, \ystart){};
\node[](mm) at (\xstart+0.9, \ystart){$\dots$};
\node[every neuron](hd2n) at (\xstart+1.3, \ystart){};

\def\ystart{-13.3}

\def\xstart{0.6}
\draw[rounded corners,dashed,fill=gray!30] (\xstart,\ystart) -- (\xstart+1.9,\ystart) -- (\xstart+1.9, \ystart+0.6) -- (\xstart, \ystart+0.6) -- cycle;

\def\xstart{-0.55}
\draw[rounded corners,dashed,fill=purple!40] (\xstart,\ystart) -- (\xstart+0.6,\ystart) -- (\xstart+0.6, \ystart+0.6) -- (\xstart, \ystart+0.6) -- cycle;

\def\ystart{-13}

\def\xstart{-0.25}
\node[every neuron](y1) at (\xstart, \ystart){$\bar{y}$};

\def\xstart{0.9}
\node[every neuron](hd31) at (\xstart, \ystart){};
\node[every neuron](hd32) at (\xstart+0.5, \ystart){};
\node[](mm) at (\xstart+0.9, \ystart){$\dots$};
\node[every neuron](hd3n) at (\xstart+1.3, \ystart){};


\draw[blue] [->] (0.3,-8.3) -- (1.55,-9.67); 

\draw[red] [->] (2.7,-8.3) -- (1.55,-9.67); 



\draw[rounded corners] [->] (-0.6, -7) -- (-1.2, -7) -- (-1.2, -9.9) -- (1.1, -9.9);

\draw[rounded corners] [->] (-1.5, -3.3) -- (-1.5, -10.1) -- (1.1, -10.1);

\draw[rounded corners] [->] (-3.4, -1.2) -- (-3.4, -8) -- (-0.6, -8);

\draw[rounded corners] [->] (-3.4, -1.45) -- (6.75, -1.45) -- (6.75, -1.7);

\draw[rounded corners] [->] (-3.4, -3) -- (-2.2, -3);

\draw[rounded corners] [->] (-3.4, -6) -- (0.3, -6) -- (0.3, -6.7);

\draw[rounded corners] [->] (7.4, -2) -- (7.6, -2) -- (7.6, -10) -- (5.95, -10);


\draw[rounded corners] [->] (0.6, -13) -- (0.05, -13);

\foreach \i in {1,2,n}
    \foreach \j in {1,2,n}{
        \draw [->] (h1\i) -- (h2\j);
        \draw [->] (h2\i) -- (h3\j);};

\foreach \i in {1,2,n}
    \foreach \j in {1,2,n}{
        \draw [->] (hd1\i) -- (hd2\j);
        \draw [->] (hd2\i) -- (hd3\j);};

\node[align=center, above] at (1.6, -2.2) {$\mathbf{z}$};
\node[align=center, above] at (5.8, -2.2) {$\mathbf{q}$};
\node[align=center, above] at (-1.5, -2.7) {$\mathbf{r}$};
\node[align=center, above] at (4.8, -3.65) {$\mathbf{e}^{G}$};
\node[align=center, above] at (3.9, -8.15) {$\hat{\mathbf{x}}$};
\node[align=center, above] at (0, -6.7) {$\mathbf{m}$};
\node[align=center, above] at (0.35, -7.7) {$\mathbf{x}$};
\node[align=center, above] at (3.6, -10.65) {$\mathbf{e}^{D}$};
\node[align=center, above] at (6.8, -6.5) {Generator};
\node[align=center, above] at (5.8, -13.5) {Critic};
\node[align=center, above] at (4, -17.9) {Message passing graph neural network};

\end{tikzpicture}

\caption{Architecture of the digital twin model. The generator receives a noise vector $\mathbf{z}$, and categorical (e.g. tissue type; $\mathbf{q}$) and numerical (e.g. age; $\mathbf{r}$) covariates, and outputs a vector of synthetic data ($\hat{\mathbf{x}}$). The critic receives data from two input streams (real, blue; and synthetic, red), a mask $\mathbf{m}$ indicating which components of the input vector are missing, and the numerical $\mathbf{r}$ and categorical $\mathbf{q}$ covariates. The critic produces an unbounded scalar $\bar{y}$ that quantifies the degree of realism of the input samples from the two input streams. The handcrafted ODE system proposed in \cite{barbiero2020computational} is used to determine a graph representation of patient's physiology. The message passing neural network updates latent node features to estimate global attributes describing the evolution of the underlying physiological system.}

\label{fig:architecture}

\end{figure}

%% file: gan_framework.tex
\begin{figure}[!h]
\centering
\newcommand\initialy{4}
\newcommand\nodeSize{0.6cm}

\tikzset{%
  every neuron/.style={
    circle,
    draw,
    fill=white,
    minimum size=\nodeSize
  },
  neuron missing/.style={
    draw=none, 
    scale=1,
    fill=none,
    text height=0cm,
    execute at begin node=\color{black}$\hdots$
  },
}

\begin{tikzpicture}[x=1.5cm, y=1.5cm, >=stealth]

\node at (0,-1)  [rectangle, rounded corners, fill=yellow!40] (prob) {$P(\mathbf{x} \text{ is  real})$};

\node at (0,0) [rectangle,draw,rounded corners, fill=blue!30] (discr) {$D(\mathbf{x})$};

\node at (-1.4,0) {Discriminator};

\node at (1.5,1) [rectangle, rounded corners, fill=gray!30] (xfake) {$\mathbf{x}_{fake}$};

\node at (-1.5,1) [rectangle, rounded corners, fill=gray!30] (xreal) {$\mathbf{x}_{real}$};

\node at (1.5,2) [rectangle,draw, rounded corners, fill=blue!30] (gen) {$G(\mathbf{z})$};

\node at (2.7,2) {Generator};

\node at (1.5,3) [rectangle, rounded corners, fill=gray!10] (prior) {$\mathbf{z} \sim p_{\mathbf{z}}$};

\node at (2.8,3) {Noise prior};

\node at (-4, 0) {};
\node at (5, 0) {};

\draw [->] (prior) -- (gen);
\draw [->] (gen) -- (xfake);
\draw [->] (xfake) -- (discr);
\draw [->] (xreal) -- (discr);
\draw [->] (discr) -- (prob);

\end{tikzpicture}
\caption[Generative Adversarial Network framework]{Generative Adversarial Network framework. The generator $G(\mathbf{z})$ receives a vector $\mathbf{z}$ sampled from a noise prior distribution $p_{\mathbf{z}}$, and generates a synthetic sample $\mathbf{x}_{fake}$. The discriminator $D(\mathbf{x})$ tries to distinguish \emph{real} samples from \emph{fake} samples, producing the probability of $\mathbf{x}$ coming from the \emph{real} data distribution. The competition between the two players drives the game and makes both players increasingly better.}

\label{gan}

\end{figure}

%% file: results_GNN.tex
\begin{figure}
\centering
\newcommand\initialy{4}
\newcommand\nodeSize{0.75cm}

\tikzset{%
  tipSquare/.tip={Circle[open]}
}

\tikzset{%
  every neuron/.style={
    circle,
    draw,
    fill=white,
    scale=0.8,
    minimum size=\nodeSize
  },
  neuron missing/.style={
    draw=none, 
    scale=0.8,
    fill=$\dots$,
    text height=0cm,
    execute at begin node=$\dots$
  },
  snake it/.style={
    decorate, decoration=snake
  }
}

\begin{tikzpicture}[x=1.5cm, y=1.5cm, >=stealth, scale=0.55, every node/.style={transform shape}, curved arrow/.style={arc arrow={to pos #1 with length 2mm and options {}}},
reversed curved arrow/.style={arc arrow={to pos #1 with length 2mm and options reversed}}]  

\node[inner sep=0pt] at (-4.4,5.5)
    {\includegraphics[width=1\textwidth]{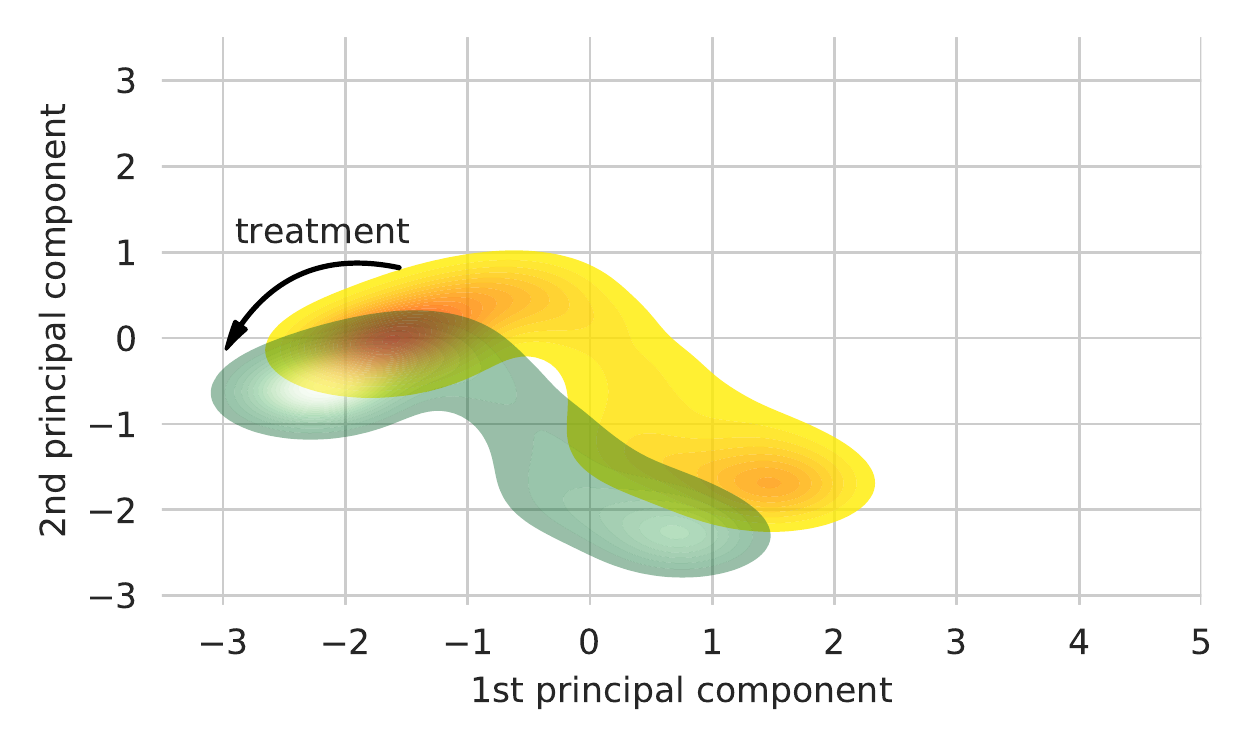}};
\node[inner sep=0pt] at (4.4,5.5)
    {\includegraphics[width=1\textwidth]{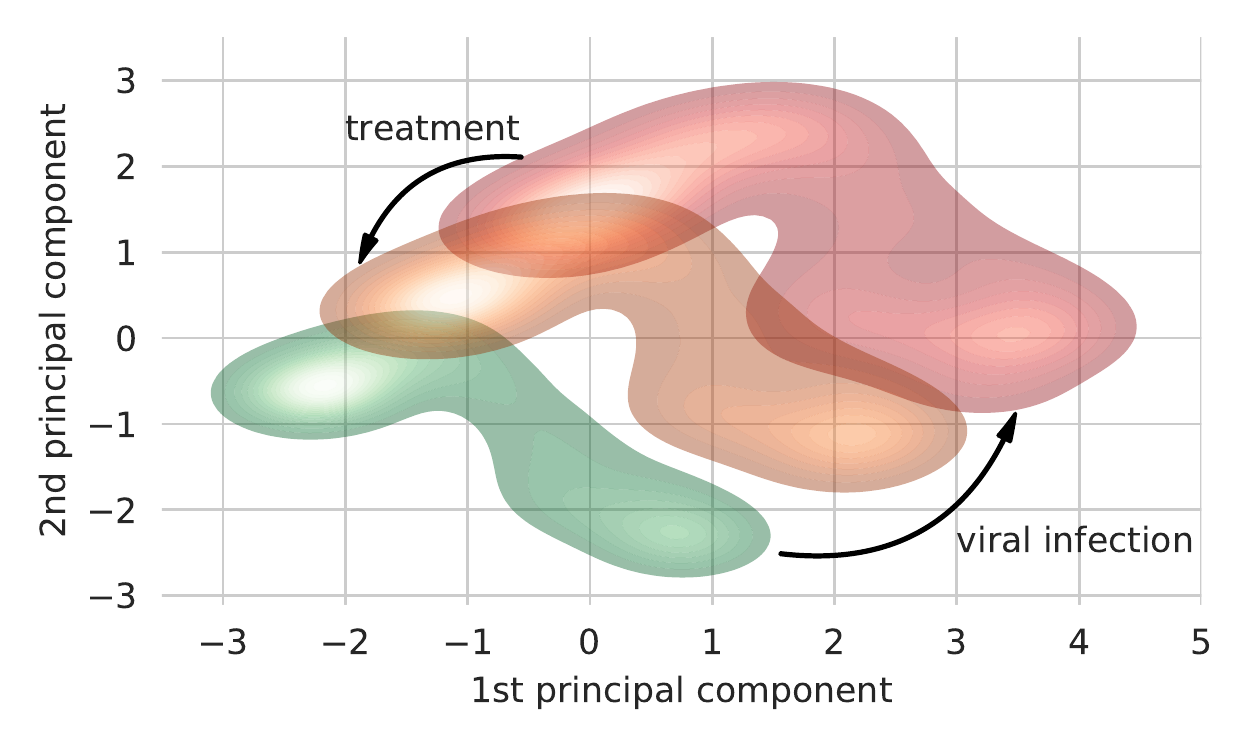}};
\node[inner sep=0pt] at (0,0)
    {\includegraphics[width=.22\textwidth]{figs/silhouette1.png}};
\node[inner sep=0pt] at (-4.4,-6)
    {\includegraphics[width=1\textwidth]{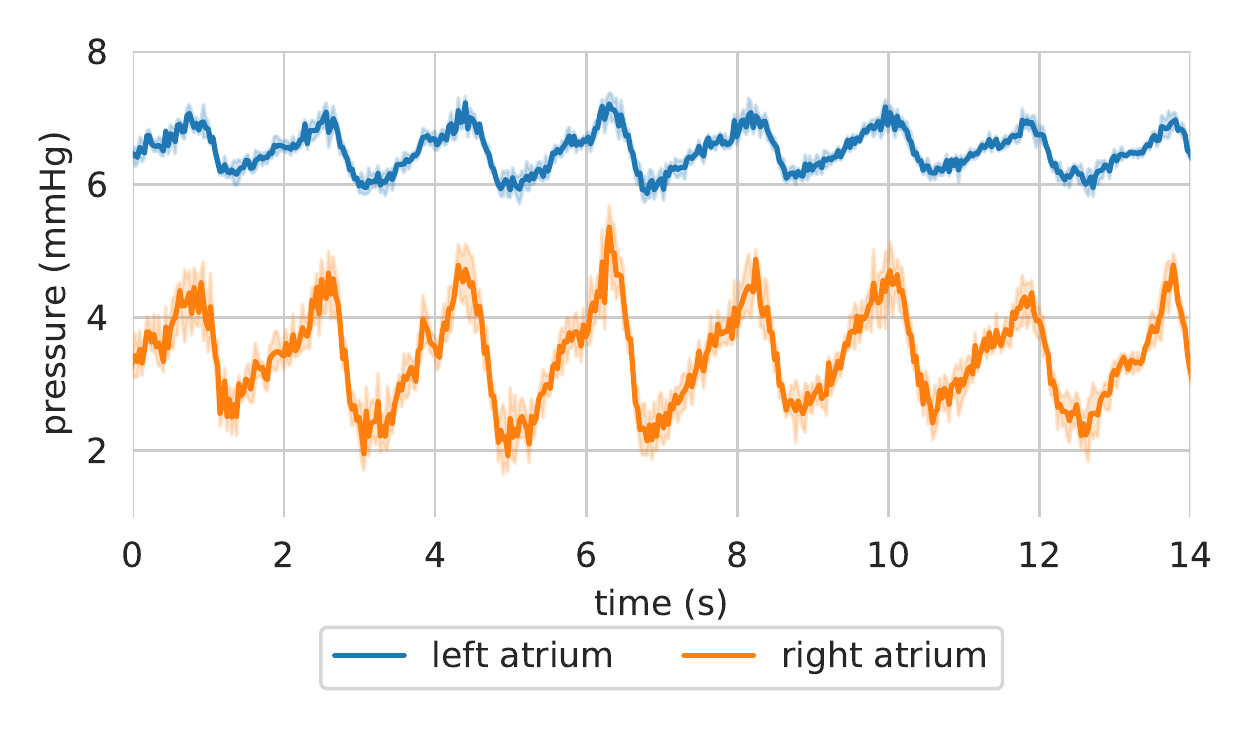}};
\node[inner sep=0pt] at (4.4,-6)
    {\includegraphics[width=1\textwidth]{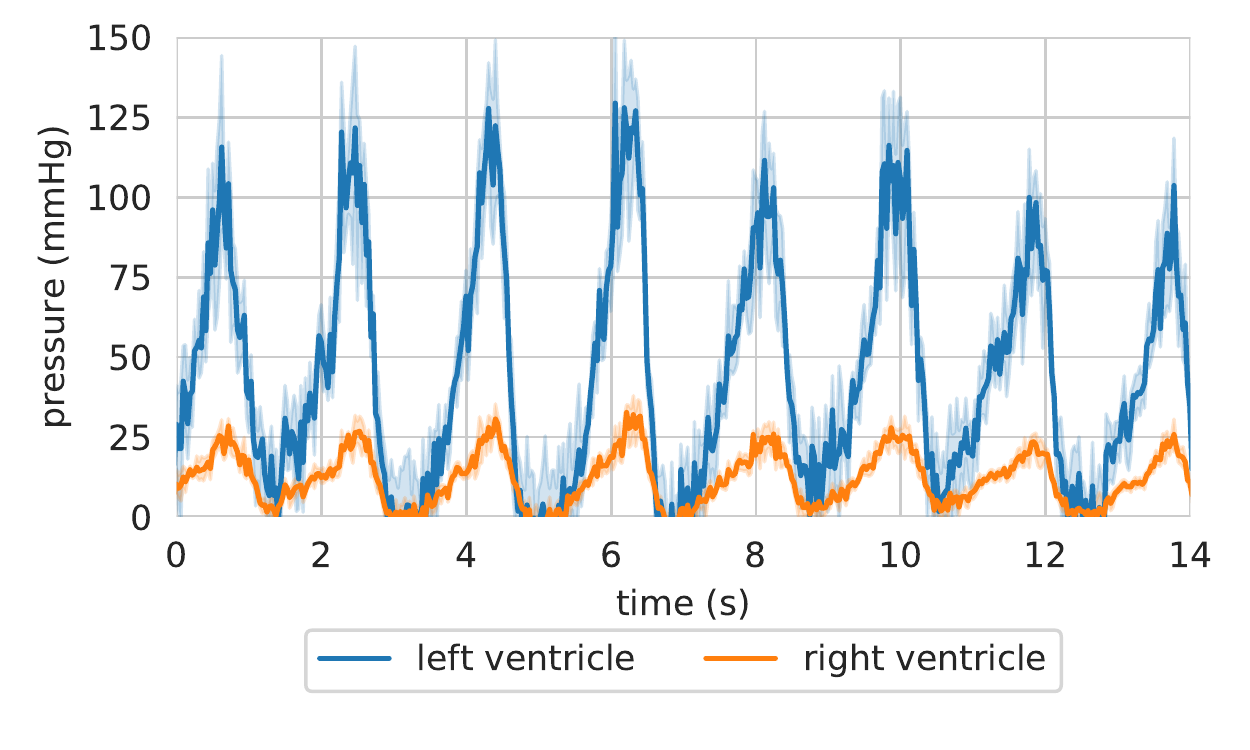}};

\draw[-{Circle[]}, black!50!red, thick] (-5,2.2) -- (-5,1.3) -- (0.2,1.3);
\draw[black!50!red, thick] [<-] (-5,2.2) -- (-5,1.3) -- (0.2,1.3);
\draw[black!50!red, thick] [<-] (5,2.2) -- (5,1.3) -- (0,1.3);

\draw[blue, densely dotted, thick] [->] (0.1,1.3) -- (-7, -1) -- (-7, -3.6);
\draw[orange, densely dotted, thick] [->] (0.1,1.3) -- (-3.3, -1) -- (-3.3, -4.8);
\draw[blue, densely dotted, thick] [->] (0.1,1.3) -- (6.8, -1) -- (6.8, -3.6);
\draw[orange, densely dotted, thick] [->] (0.1,1.3) -- (3.2, -1) -- (3.2, -6.3);

\end{tikzpicture}

\caption{Two clinical case studies represented in a projected heart phase-space. The first case study (left figure) shows the effect of a therapeutic intervention comprising an increased physical exercise, a reduced amount of calorie intake, and the subscription of a daily dosage of Benazepril (5mg). The second simulation (right figure) shows the long-term impact on blood pressure of an untreated SARS-CoV-2 infection (red density) and the effects of a therapy including both Benazepril (5 mg/day) and intra venous injection of heparin (5000 U/ml) (orange density). [\textbf{top figures}]
Bundle of predicted trajectories can be visualised and monitored in real time in order to investigate patterns in the time domain. The simulation shows blood pressure in heart chambers starting from healthy state conditions. Error bands represent 95\% CI. [\textbf{bottom figures}]}

\label{fig:gnn_results}

\end{figure}

%% file: results_gene_associations.tex
\clearpage

\begin{landscape}

\begin{figure}
\centering
\newcommand\initialy{4}
\newcommand\nodeSize{0.75cm}

\tikzset{%
  tipSquare/.tip={Circle[open]}
}

\tikzset{%
  every neuron/.style={
    circle,
    draw,
    fill=white,
    scale=0.8,
    minimum size=\nodeSize
  },
  neuron missing/.style={
    draw=none, 
    scale=0.8,
    fill=$\dots$,
    text height=0cm,
    execute at begin node=$\dots$
  },
  snake it/.style={
    decorate, decoration=snake
  }
}

\begin{tikzpicture}[x=1.5cm, y=1.5cm, >=stealth, scale=0.54, every node/.style={transform shape}, curved arrow/.style={arc arrow={to pos #1 with length 2mm and options {}}},
reversed curved arrow/.style={arc arrow={to pos #1 with length 2mm and options reversed}}, scale=1]  

\node[inner sep=0pt] at (0,-0.05)
    {\includegraphics[width=.8\textwidth, height=0.65\textwidth]{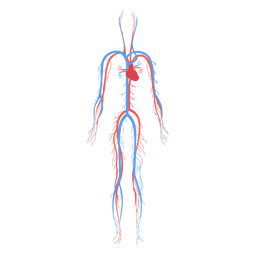}};
\node[inner sep=0pt] at (0,0)
    {\includegraphics[width=.3\textwidth]{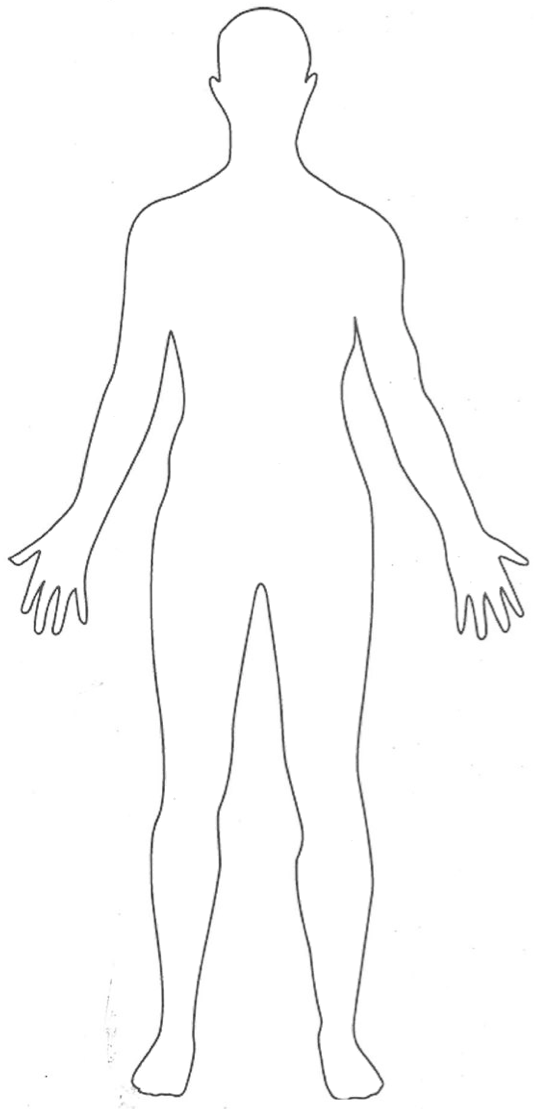}};
    
\node[inner sep=0pt] at (-7, 5.4)
    {\includegraphics[scale=0.5]{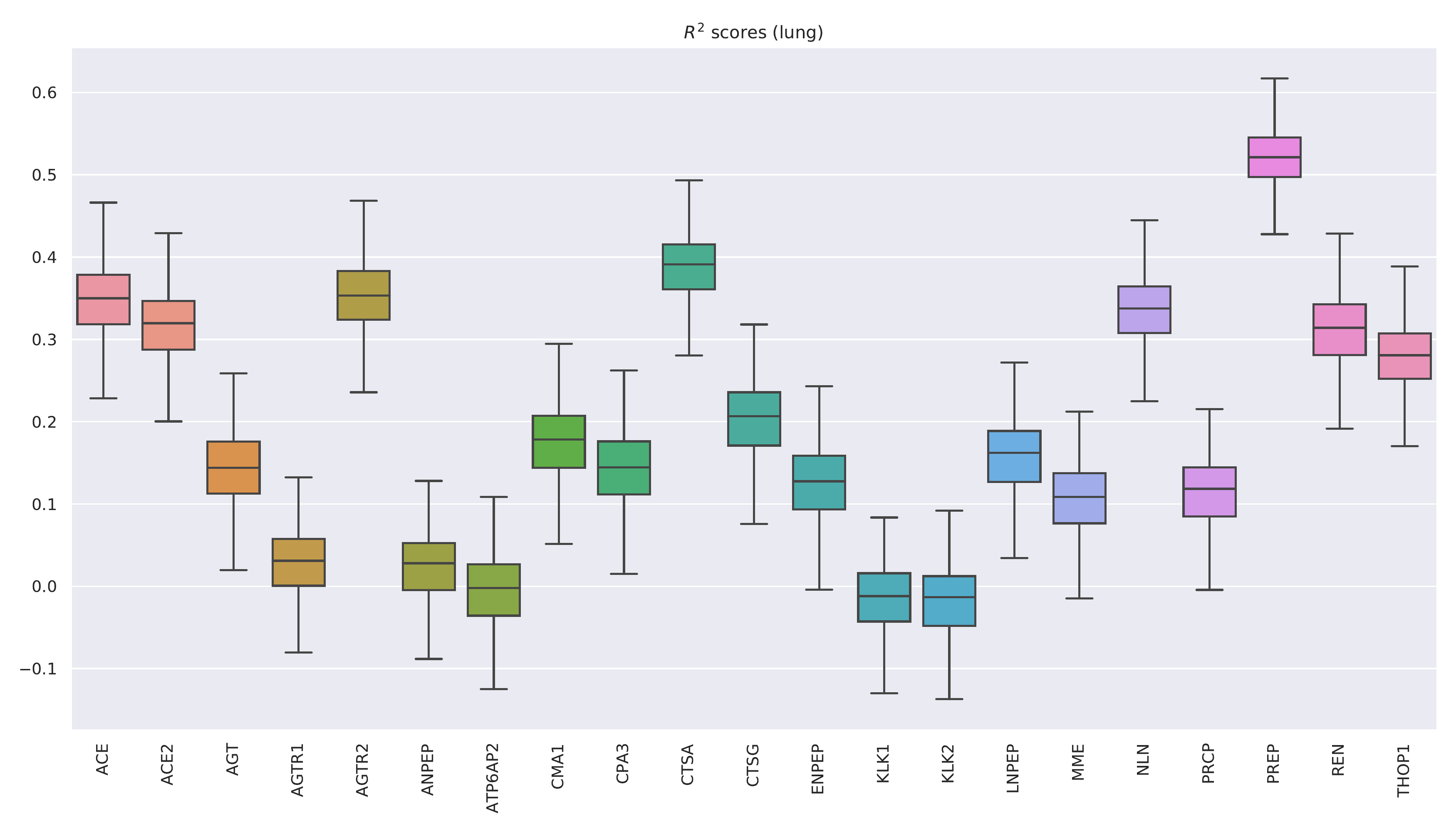}};
\node[inner sep=0pt] at (6.7, 5.4)
    {\includegraphics[scale=0.5]{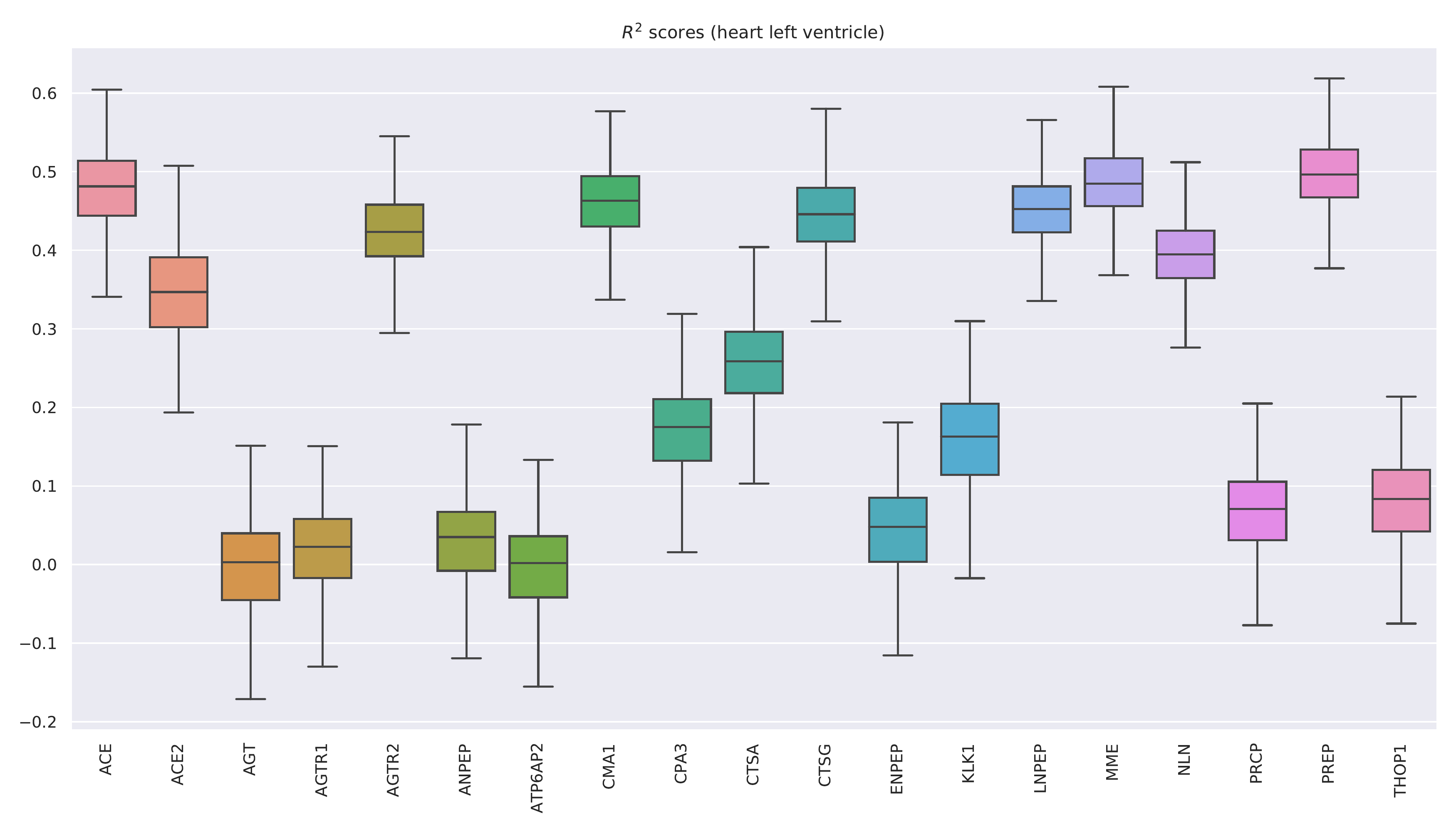}};

\node[inner sep=0pt] at (-7,-4.5)
    {\includegraphics[scale=0.5]{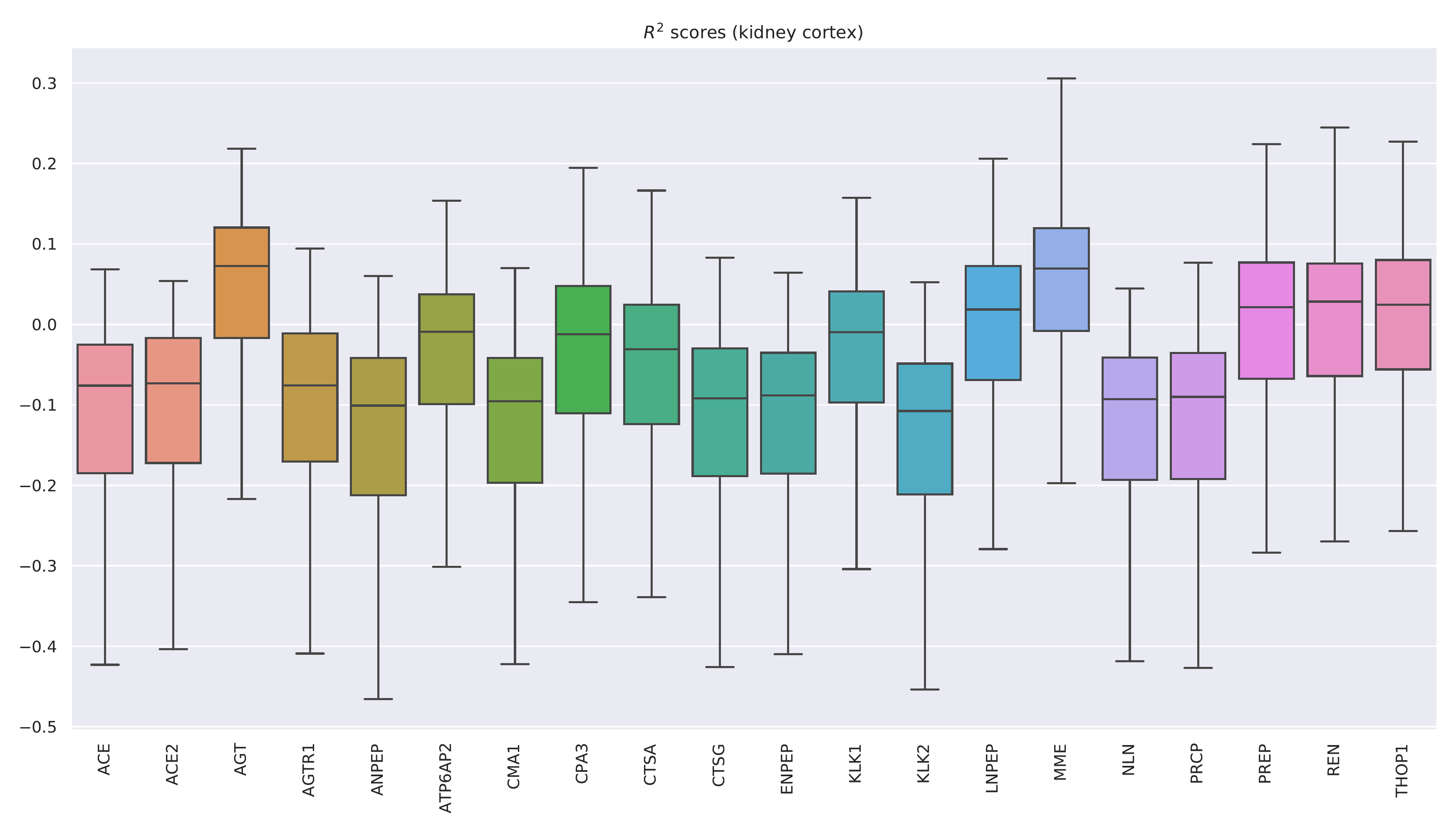}};
    
\node[inner sep=0pt] at (6.7,-4.5)
    {\includegraphics[scale=0.5]{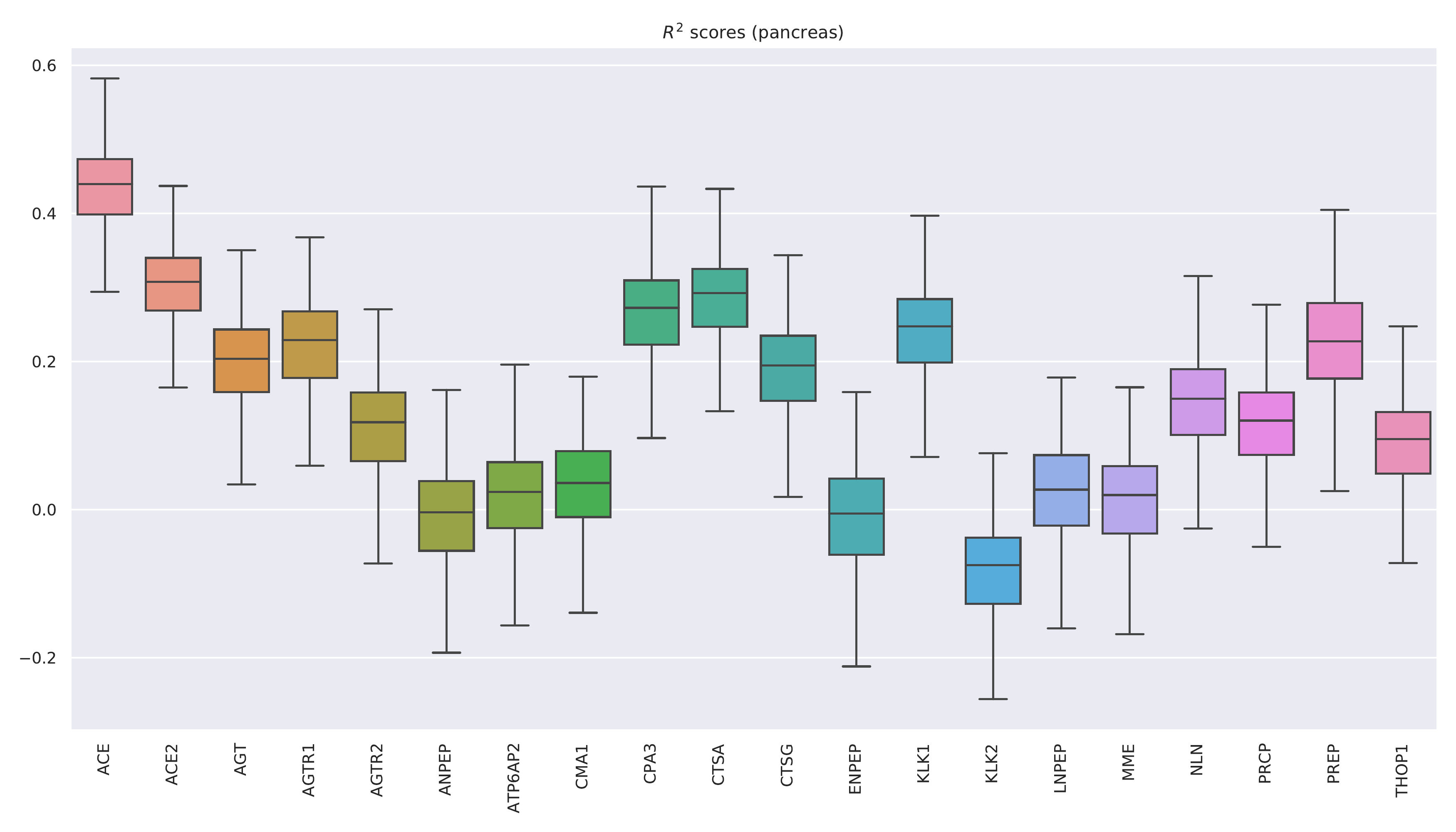}};

\draw[blue, densely dotted, thick] [->] (-0.3,1.5) -- (-7, 1.5) -- (-7, 2.2);
\draw[-{Circle[]}, black!50!red, thick] (-0.3,1.6);
\draw[orange, densely dotted, thick] [->] (0.2,1.4) -- (7, 1.4) -- (7, 2.2);
\draw[-{Circle[]}, black!50!red, thick] (0.2,1.5);
\draw[red, densely dotted, thick] [->] (-0.3, 1) -- (-7, 1) -- (-7, -1.2);
\draw[-{Circle[]}, black!50!red, thick] (-0.3, 1.1);
\draw[black!50!green, densely dotted, thick] [->] (0.,1) -- (7, 1) -- (7, -1.2);
\draw[-{Circle[]}, black!50!red, thick] (0, 1.1);

\end{tikzpicture}

\caption{Bootstrapped $R^2$ scores for genes involved in the renin-angiotensin system for lung, heart (left ventricle), kidney (cortex), and pancreas. The input variables are the expressions of genes in whole blood belonging to the chemokine, TNF, and TFG-$\beta$ pathways.}

\label{fig:association_results}

\end{figure}

\end{landscape}